\documentclass[10pt,twocolumn,letterpaper]{article}

\usepackage{wacv}              

\usepackage{graphicx}
\usepackage{amsmath}
\usepackage{amssymb}
\usepackage{booktabs}
\usepackage{newfloat}
\usepackage{listings}
\usepackage{subcaption}
\usepackage{makecell}
\usepackage{cuted}
\usepackage{colortbl}
\usepackage{multirow}
\usepackage{booktabs}
\usepackage{mathtools}
\usepackage{bm, amsthm, amssymb}
\usepackage{filecontents}
\usepackage{lipsum}
\usepackage{etoolbox}
\usepackage{amsmath,amsfonts}
\usepackage{algorithmic}
\usepackage{algorithm}
\usepackage{array}
\usepackage{textcomp}
\usepackage{stfloats}
\usepackage{url}
\usepackage{verbatim}
\usepackage{graphicx}
\usepackage{cite}

\usepackage[pagebackref,breaklinks,colorlinks]{hyperref}
\usepackage{cuted}

\usepackage[capitalize]{cleveref}
\crefname{section}{Sec.}{Secs.}
\Crefname{section}{Section}{Sections}
\Crefname{table}{Table}{Tables}
\crefname{table}{Tab.}{Tabs.}

\def\wacvPaperID{2164} 
\def\confName{WACV}
\def\confYear{2025}

\begin{document}

\title{WINE: Wavelet-Guided GAN Inversion and Editing for High-Fidelity Refinement}

\author{
    Chaewon Kim$^{*}$\thanks{Co-first Author}\thanks{This research was done at Klleon AI Research center.}\\
    KRAFTON \\
    \texttt{hkchae96@gmail.com}
    \and
    Seung Jun Moon$^{*}$ \\
    Klleon AI Research \\
    \texttt{june1034@gmail.com}
    \and
    Gyeong-Moon Park\thanks{Corresponding Author.} \\
    Kyung Hee University \\
    \texttt{gmpark@khu.ac.kr}
}

\maketitle

\begin{strip}
\centering
\vspace{-1.9cm}
\includegraphics[width=\textwidth]{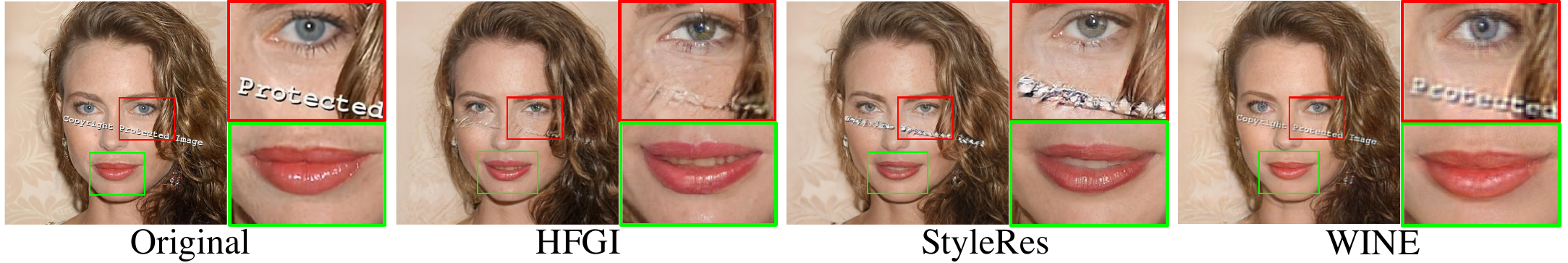}
\captionof{figure}{\textbf{
Comparison of Recent GAN Inversion Models.}
We evaluate the recent high-fidelity GAN inversion models with the inversion of an intricate image. Specific regions demanding meticulous preservation of details, such as letters, lip shape, and eye pupil, are closely examined. Even with the high-rate inversion via residual learning, existing baselines encounter difficulties in adequately restoring these nuanced details. In contrast, our newly introduced WINE method excels in the robust preservation of such intricate details.}
\label{fig:front}
\end{strip}

\begin{abstract}

Recent advanced GAN inversion models aim to convey high-fidelity information from original images to generators through methods using generator tuning or high-dimensional feature learning.
Despite these efforts, accurately reconstructing image-specific details remains as a challenge due to the inherent limitations both in terms of training and structural aspects, leading to a bias towards low-frequency information.
In this paper, we look into the widely used pixel loss in GAN inversion, revealing its predominant focus on the reconstruction of low-frequency features. We then propose WINE, a \textbf{W}avelet-guided GAN \textbf{I}nversion a\textbf{N}d \textbf{E}diting model, which transfers the high-frequency information through wavelet
coefficients via newly proposed wavelet loss and wavelet fusion scheme. Notably, WINE is the first attempt to interpret GAN inversion in the frequency domain. Our experimental results showcase the precision of WINE in preserving high-frequency details and enhancing image quality. 
Even in editing scenarios, WINE outperforms existing state-of-the-art GAN inversion models with a fine balance between editability and reconstruction quality. 
\end{abstract}

\section{Introduction}
\begin{figure*}[t]
    \centering
    \begin{subfigure}{0.22\textwidth}
        \centering
        \includegraphics[width=\textwidth]{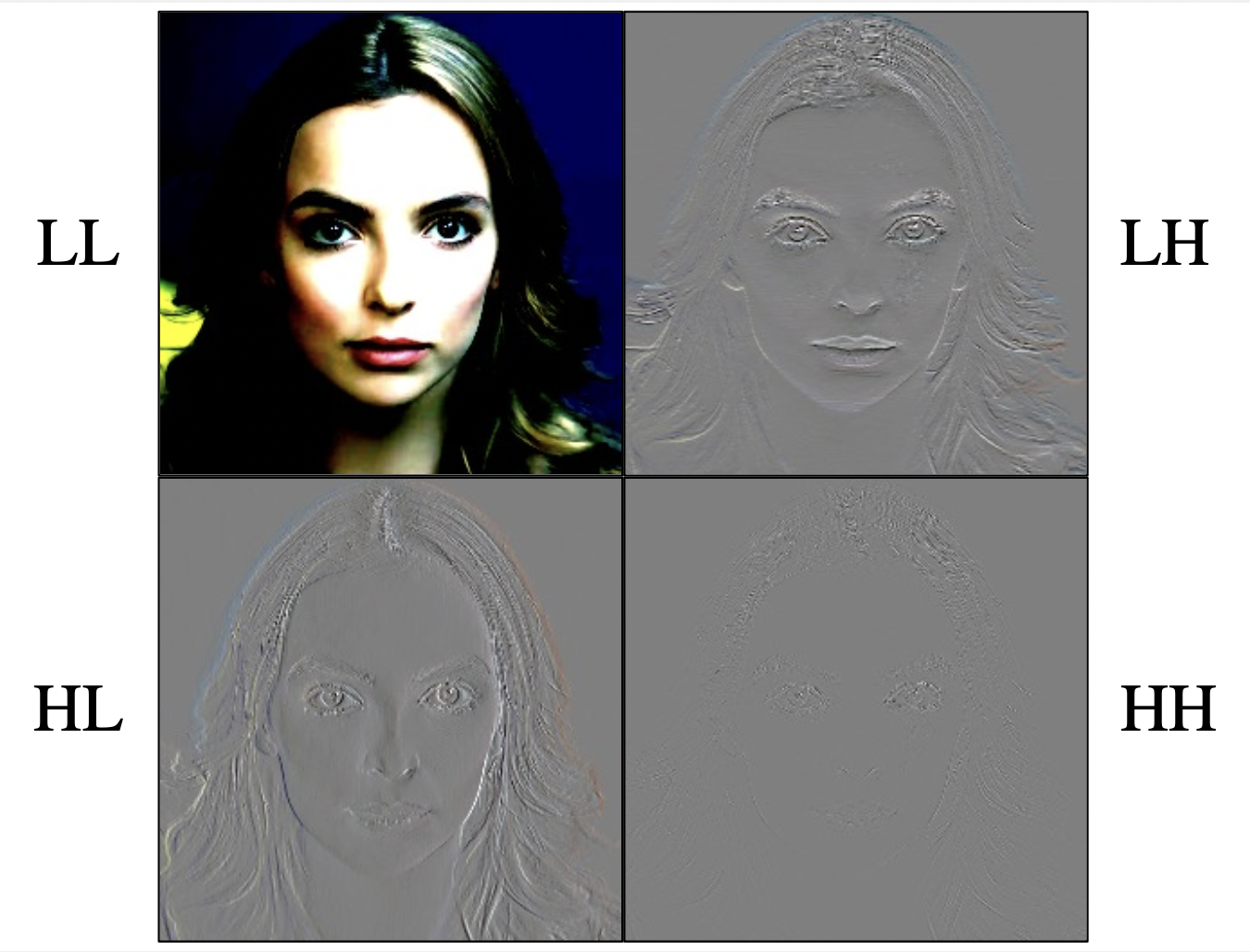}
        \caption{\scriptsize{Wavelet transform}}
        \label{fig:dwt}
    \end{subfigure}
    \hfill
    \begin{subfigure}{0.33\textwidth}
        \centering
        \includegraphics[width=\textwidth]{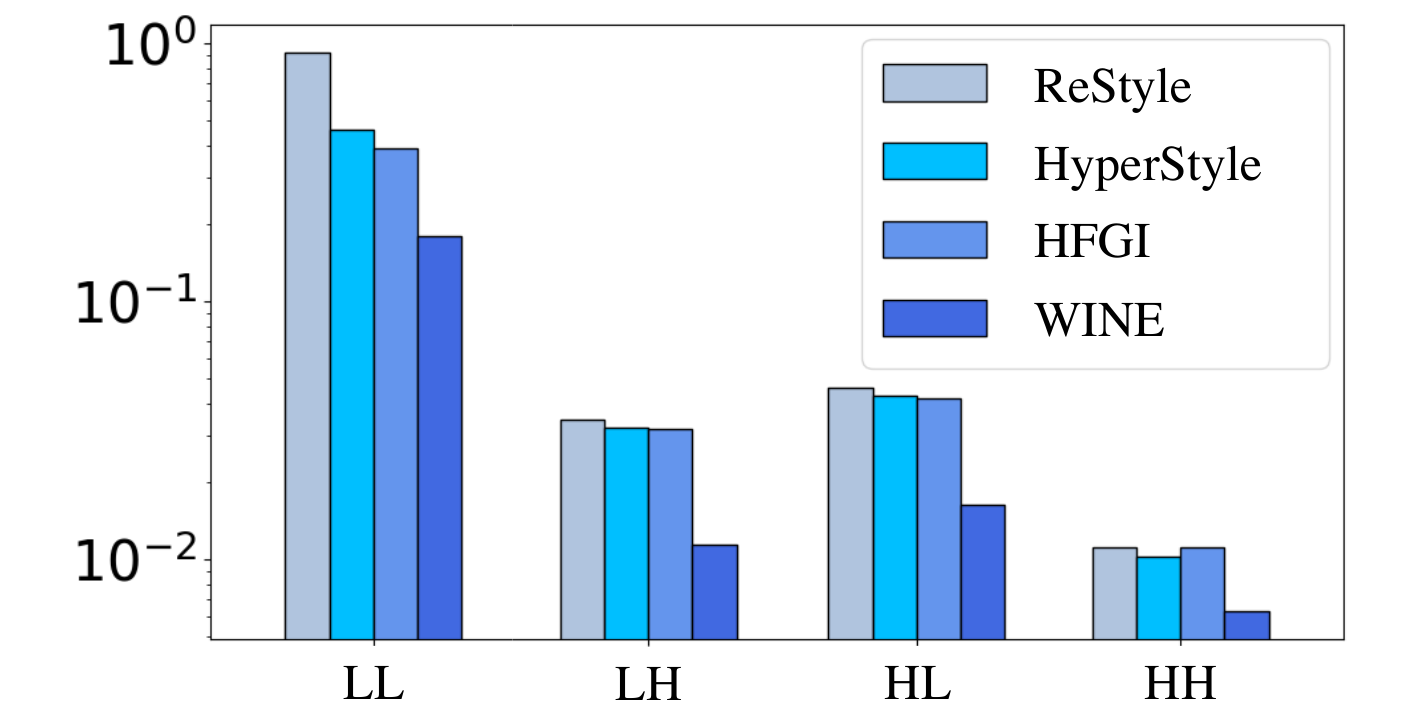}
        \caption{\scriptsize{$L_{2}$ of wavelet coefficients.}}
        \label{fig:comp_l2}
    \end{subfigure}
    \hfill
    \begin{subfigure}{0.32\textwidth}
        \centering
        \includegraphics[width=\textwidth]{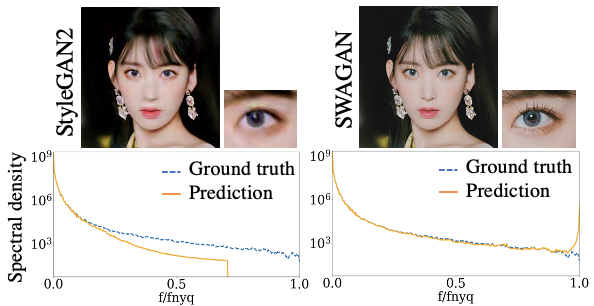}
        \caption{\scriptsize{Reconstruction quality of generators.}}
        \label{fig:comp_frequency}
    \end{subfigure}
    \caption{
    (a) \textbf{Wavelet Transform}. We plot the wavelet coefficients by each filter at $1^{st}$ wavelet decomposition.
    The gray color denotes the zero value.
    Coefficients from $LH$, $HL$, and $HH$, have significantly high sparsity than the coefficient from $LL$.
    (b) \textbf{Comparison of $\mathcal{L}_{2}$ from Each Filter between Baseline Methods}. We plot the average $\mathcal{L}_{2}$ of each wavelet coefficient between CelebA-HQ test images and corresponding inverted images by various state-of-the-art inversion models.
    Due to the significant gap between $\mathcal{L}_{2,LL}$ and the rest ($\sim 30 \times$ in linear scale), we display the losses with the logarithmic scale.    (c)  \textbf{Comparison of Image Reconstruction Quality between Backbone Generators}. We compared the single-image reconstruction ability of StyleGAN2 and SWAGAN in both visual and spectral aspects. As SWAGAN generates images in the spatial frequency domain, the generated image preserves high-frequency information.}
    \label{fig:variance}
\end{figure*}

\noindent Following the impressive generation capabilities of unconditional Generative Adversarial Networks (GANs) \cite{goodfellow2020generative}, inversion models of GAN have witnessed significant advancements. These improvements manifest in terms of robust reconstruction and disentangled image attribute editing \cite{abdal2019image2stylegan, alaluf2021restyle, alaluf2021hyperstyle, richardson2021encoding, tov2021designing, wang2022high, pehlivan2023styleres, moon2022interestyle, parmar2022spatially, liu2023survey, liu2023delving, katsumata2023revisiting}.
Due to the property of unconditional GAN, inversion models should find adequate latent within the latent space of GAN to edit given images.
The early GAN inversion models mostly rely on per-image optimization \cite{abdal2019image2stylegan, abdal2020image2stylegan++, zhu2020domain}, which is extremely time-consuming.
For real-time inference, the encoder-based method becomes prevalent \cite{alaluf2021restyle, moon2022interestyle, richardson2021encoding, tov2021designing, pehlivan2023styleres, alaluf2021hyperstyle}, which trains an encoder that returns the corresponding GAN latent of an input image.
The acquired latent from the encoder is desired to reproduce the input image as closely as possible.

However, the encoder needs to compress the image into a small dimension, \ie, low-rate inversion.
For the case of StyleGAN2 \cite{karras2020analyzing}, in order to encode an image $X \in \mathbb{R}^{1024\times1024\times3}$ the encoder returns the corresponding latent $w \in \mathbb{R}^{18\times512}$, which is extremely smaller than the original image dimension (about 0.3\%).
Due to the Information Bottleneck theory \cite{tishby2015deep, wang2022high, pehlivan2023styleres}, encoding information into a small tensor causes severe information loss, and it aggravates the high-frequency image details.

To overcome this shortage, recent GAN inversion models propose new directions, such as fine-tuning the generator \cite{alaluf2021hyperstyle, roich2021pivotal} or manipulating the intermediate feature of GAN \cite{wang2022high} to deliver more information using higher dimensional features than latents, \ie, high-rate inversion.
However, the results of high-rate inversions are still imperfect.
Figure \ref{fig:front} shows the inversion results of high-rate inversion models, HFGI \cite{wang2022high} and StyleRes \cite{pehlivan2023styleres}.
Though both models generally preserve coarse features, details are distorted, \eg, boundaries of letters and details of the face.

Consequently, we hypothesize that the lower distortion of high-rate inversion models compared to the state-of-the-art low-rate inversion model (ReStyle \cite{alaluf2021restyle}) is owing to reducing distortion on low-frequency sub-bands.
To explicitly check the distortion per frequency sub-band, we adopt a wavelet transform, which enables the use of both frequency and spatial information.
The wavelet transform returns a total of four coefficients by passing filters: a low-pass filter set $\mathbb{F}_{l}=\{LL\}$ and a high-pass filter set $\mathbb{F}_{h}=\{LH, HL, HH\}$.
In Figure \ref{fig:dwt}, we visualize the coefficients obtained by each filter.
In Figure \ref{fig:comp_l2}, we compare $\mathcal{L}_{2}$ between the coefficients of ground truth images and inverted images, yielded by a filter $f$, \ie, $\mathcal{L}_{2,f}$.
While the high-rate inversion models apparently decrease $\mathcal{L}_{2,f}$ for $f \in \mathbb{F}_{l}$, they marginally decrease or even increase $\mathcal{L}_{2,f}$ for $f \in \mathbb{F}_{h}$, compared to ReStyle.
In light of this observation, we can argue that the high-rate existing methods can decrease distortion on the low-frequency sub-band, but are not effective for decreasing distortion on the high-frequency sub-band.

\textbf{Contributions.}
We propose WINE, a new \textbf{W}avelet-guided GAN \textbf{I}nversion a\textbf{N}d \textbf{E}diting model.
WINE effectively preserves high-frequency details on both inversion and editing scenarios by explicitly handling high-frequency features.
To the best of our knowledge, WINE is the first attempt to interpret GAN inversion in the frequency domain.
Especially, we propose two novel terms: (\lowercase{\romannumeral1}) \textit{wavelet loss} and (\lowercase{\romannumeral2}) \textit{wavelet fusion}.
First, (\lowercase{\romannumeral1}) amplifies the loss of the high-frequency sub-band by using the wavelet coefficients from $f\in\mathbb{F}_{h}$.
By using wavelet loss at training, WINE is proficient in reconstructing high-frequency details.
For the explicit comparison with the previously used loss term, \ie, $\mathcal{L}_{2}$, we quantitatively show the low-frequency bias of $\mathcal{L}_{2}$, compared to the wavelet loss.
Second, (\lowercase{\romannumeral2}) transfers the high-frequency features directly to the wavelet coefficients of the reconstructed image.
Due to the wavelet upsampling structure of SWAGAN \cite{gal2021swagan}, we can explicitly manipulate the wavelet coefficients during the hierarchical upsampling.

We demonstrate that WINE shows outstanding results, compared to the existing state-of-the-art GAN inversion models \cite{alaluf2021hyperstyle, wang2022high, pehlivan2023styleres}.
WINE achieves the lowest distortion among the existing GAN inversion models.
Moreover, qualitative results show the robust preservation of image-wise details of our model, both on the inversion and editing scenarios via InterFaceGAN \cite{shen2020interfacegan} and StyleCLIP \cite{patashnik2021styleclip}.
Finally, we elaborately show the ablation results and prove that each of our proposed methods is indeed effective.
\section{Related Works}

\subsection{Wavelet Transform}

\noindent Wavelet transform provides information on both frequency and spatiality \cite{daubechies1990wavelet}, which are crucial in the image domain.
The most widely used wavelet transform in deep learning-based image processing is the Haar wavelet transform, which contains the following four simple filters:

\begin{align*}
& LL=
\begin{bmatrix}
1 & 1  \\
1 & 1
\end{bmatrix}
,
LH=
\begin{bmatrix}
-1 & -1  \\
1 & 1
\end{bmatrix}
, \\
& HL=
\begin{bmatrix}
-1 & 1  \\
-1 & 1
\end{bmatrix}
,
HH=
\begin{bmatrix}
1 & -1  \\
-1 & 1
\end{bmatrix}.
\end{align*}

Since the Haar wavelet transform enables reconstruction without information loss via inverse wavelet transform \cite{yoo2019photorealistic}, it is widely used in image reconstruction-related tasks, \eg, super-resolution \cite{huang2017wavelet, liu2018multi} and photo-realistic style transfer \cite{yoo2019photorealistic}.
In GAN inversion, image-wise details that cannot be generated via generator, \eg, StyleGAN \cite{karras2019style, karras2020analyzing}, should be transferred without information loss.
Consequently, we argue that the Haar wavelet transform is appropriate for GAN inversion.
Our method is the first approach to combine the wavelet transform in GAN inversion.

\subsection{Frequency Bias of Generative Models}

\noindent Recent works address the spectral bias in GANs \cite{Dzanic2019fourier, Liu2020fake, wang2019cnngenerated}, where the training is biased to learn the low-frequency distribution whilst struggling to learn the high-frequency counterpart.
GANs suffer from generating high-frequency features due to incompatible upsampling operations in the pixel domain \cite{Khayatkhoei2022spatialbias, schwarz2021freqbias}.
In order to alleviate the spectral distortions, prior works \cite{durall2020upconv, jiang2021focal, schwarz2021freqbias} propose a spectral regularization loss term to match the high-frequency distribution.
However, images generated via spectral loss contain undesirable high-frequency noises to coercively match the spectrum density, eventually degrading the image quality (see appendix for experimental results).

To overcome these limitations, SWAGAN \cite{gal2021swagan} introduces a wavelet-guided StyleGAN2 via generating images with the hierarchical growth of predicted wavelet coefficients. This expansion to the wavelet domain maintains the spatial frequency information, ensuring the accurate approximation of spectral density. 
In Figure \ref{fig:comp_frequency} we show the single-image reconstruction quality of StyleGAN2 and SWAGAN using a latent optimization \cite{karras2019style}. 
Spectral density plots \cite{schwarz2021freqbias} of generated and ground truth images for both architectures indicate the effectiveness of the wavelet reconstruction in preserving frequency information, particularly in the high-frequency range.
Motivated by these, we extend the application of wavelet-guided expansion beyond the generator and incorporate it into the entire GAN inversion framework.

\begin{figure*}[!t]
\begin{center}
\includegraphics[width=1\textwidth]{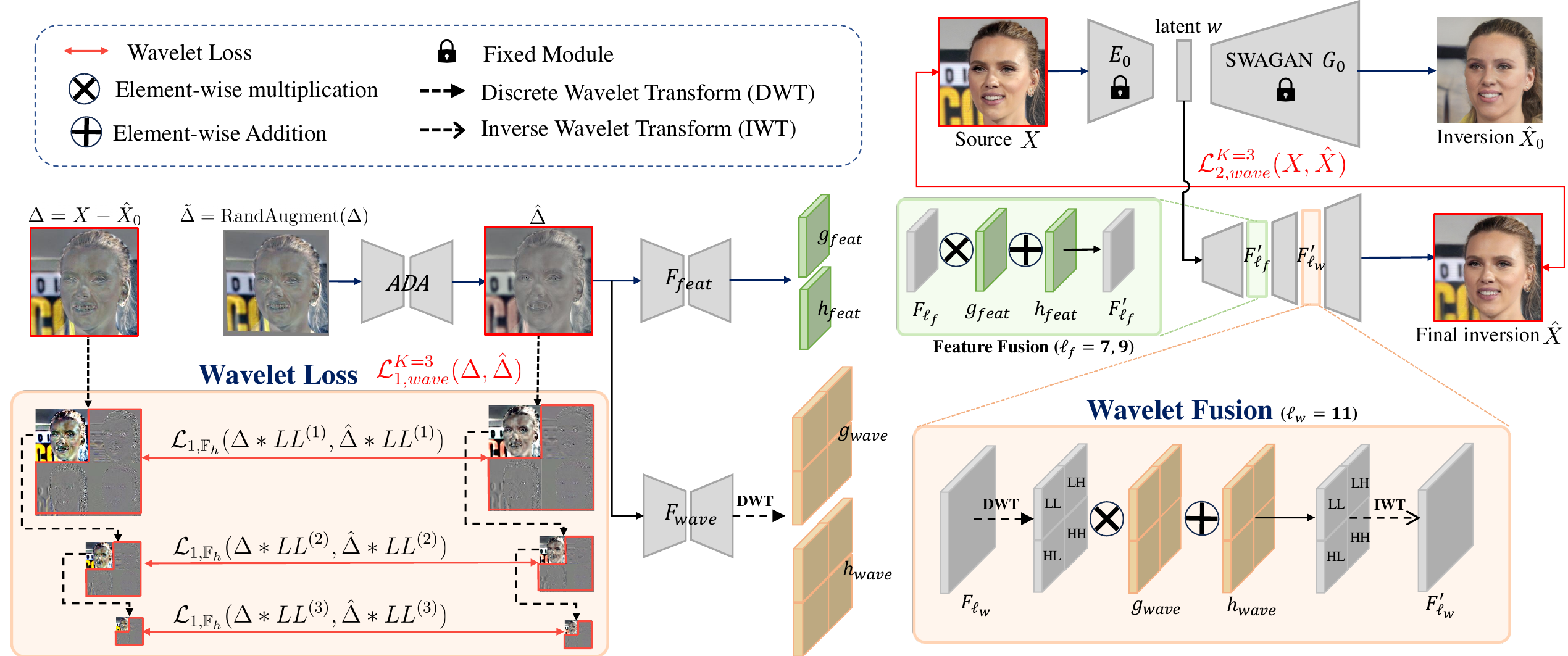}
\end{center}
\caption{
\textbf{Training Scheme of WINE.}
Given a pre-trained encoder $E_0$ and generator $G_0$, we can obtain an initial inverted image $\hat{X}_0 $. The residual $\Delta$ contains high-fidelity details that $\hat{X}_0$ misses.
The model leverages a trainable Adaptive Distortion Alignment ($ADA$) module to align the residual, which should ultimately be in alignment with $\hat{X}_0$ or the edited image $\hat{X}_0^{edit} $ at inference.
From the aligned $\hat{\Delta}$, we can replenish the missing high-fidelity information with the two fusion modules $F_{feat}$ and $F_{wave}$.
Fusion with each output is operated in the feature and frequency domain in separate intermediate layers.
The final inversion result $\hat{X}$ contains rich information without the loss of high-frequency components.
Note that $ADA$, $F_{feat}$, and $F_{wave}$ are all jointly trained, while $E_0$ and $G_0$ are frozen.
}
\vspace{-0.3cm}
\label{fig:architecture}
\end{figure*}

\subsection{High-Rate GAN Inversion}
\label{rw:3}
\noindent The prevailing high-rate GAN inversion methods include generator tuning and the learning of residual features missing from low-rate inversion.
First, generator tuning is proposed by Pivotal Tuning \cite{roich2021pivotal}, which fine-tunes the generator to lower distortion for input images.
Since Pivotal Tuning needs extra training for every new input, it is extremely time-consuming.
HyperStyle \cite{alaluf2021hyperstyle} enables generator tuning without training, by using HyperNetwork \cite{ha2016hypernetworks}.
Though HyperStyle effectively lowers distortion compared to low-rate inversion methods, it still encounters the frequency bias.
Second, the concept of adapting the residual features to StyleGAN layers is proposed by HFGI \cite{wang2022high}.
HFGI calculates the missing information from low-rate inversion and extracts feature vectors via a consultant encoder, subsequently fused to the original StyleGAN feature.
StyleRes \cite{pehlivan2023styleres} learns the residual features in higher latent codes and transforms the feature to adapt to manipulations in latent codes for robust editing.
However, both methods rely on the low-frequency biased loss even when training the image-specific delivery module. Also, despite the incorporation of residual information, the inherent bias of the StyleGAN is still inadequate for the effective reconstruction of details, again resulting in information leakage.

\section{Method}

\noindent In this section, we propose an effective Wavelet-guided GAN Inversion and Editing method, named WINE.
We first introduce the notation and the architecture of WINE.
We then further elucidate the low-frequency bias in $\mathcal{L}_{1}$ and $\mathcal{L}_{2}$ in the aspect of wavelet transform.
To settle it, we propose a novel loss term, named \textit{wavelet loss}, which explicitly focuses on high-frequency features.
Lastly, we point out the limitation of existing feature fusion and propose \textit{wavelet fusion}, which robustly transfers high-frequency features.
By using both, WINE finally becomes capable of both transferring and manipulating high-frequency information.

\subsection{Notation and Architecture}
\label{sec:method1}

\noindent Figure~\ref{fig:architecture} shows our overall architecture.
Since the goal of WINE is to retain high-frequency details of an image $X$, we design our model to be two-stage; na\"ive inversion and addition of image-wise details.

First, the pre-trained SWAGAN $G_0$ and its pre-trained encoder $E_0$\footnote{We used $e4e$\cite{tov2021designing} for $E_{0}$.} can obtain low-rate latent, $w=E_0(X)$, and the corresponding na\"ive inversion $\hat{X}_0 = G_0(w)$.
Due to the inherent limitation of low-rate inversion, $\hat{X}_0$ misses image details, denoted by $\Delta = X-\hat{X}_0$.
$\Delta$ should adapt to the alignment of the edited image $\hat{X}_0^{edit}$ at the editing scenario.
Since $\hat{X}_0^{edit}$ might have distorted alignment, \eg, a varied face angle, eye location, etc., we train the Adaptive Distortion Alignment \cite{wang2022high} ($ADA$) module that re-aligns $\Delta$ to fit the alignment of $\hat{X}_0^{edit}$.
At the training phase, due to the absence of edited images preserving image-wise details, we impose a self-supervised task by deliberately making a misalignment via a random distorted map to $\Delta$, $\tilde{\Delta}=\text{RandAugment}(\Delta)$ \cite{wang2022high}.
The purpose of $ADA$ is to minimize the discrepancy between output $\hat{\Delta}=ADA(\hat{X}_0, \tilde{\Delta})$ and $\Delta$. 
Note that we use our proposed wavelet loss (see section \ref{sec:method2}) to reconstruct the high-frequency information.
 
WINE combines $\hat{\Delta}$ using the low-rate latent $w$ with feature fusion, $F_{feat}$ \cite{wang2022high}, and our proposed method, wavelet fusion, $F_{wave}$.
Both $F_{feat}$ and $F_{wave}$ follow the linear gated scheme to filter out redundant information, which returns the pairs (\textit{scale, shift}), \ie, $(g_{feat}, h_{feat}) = F_{feat}(\hat{\Delta})$ and $(g_{wave}, h_{wave}) = F_{wave}(\hat{\Delta})$, respectively.
The pairs of $g$ and $h$ adaptively merge desired information from $w$ and $\hat{\Delta}$.
Refer to Section \ref{sec:method3} for details of fusion processes.

\subsection{Wavelet Loss}
\label{sec:method2}
\noindent Previous works \cite{rahaman2019spectral, schwarz2021freqbias} point out that $\mathcal{L}_{2}$, the most widely used loss term for comparing the ground truth and the reconstructed image in the GAN inversion field, is biased towards the low-frequency sub-band.
Since we utilize the wavelet for debiasing, we reinterpret the frequency bias in the aspect of the wavelet.
Using the proof in the appendix, $\mathcal{L}_{2}$ can be indicated as a sum of $\mathcal{L}_{2}$ for each wavelet coefficient:
\begin{align}
\nonumber
& \mathcal{L}_{2}(I_{1},I_{2}) = \sum_{f\in\mathbb{F}_{l}\cup\mathbb{F}_{h}}\lambda_{f}\mathcal{L}_{2,f}(I_{1},I_{2}) \\
& \text{holds when}\, \lambda_{f}=1, \forall f\in \mathbb{F}_{l} \cup \mathbb{F}_{h},
\end{align}
where $I_{1}$ and $I_{2}$ are arbitrary image tensors.

Since $\mathcal{L}_{2}$ reflects $\mathcal{L}_{2,f\in\mathbb{F}_{l}\cup \mathbb{F}_{h}}$ with the same weight, it seems fair without frequency bias.
However, as shown in Figure \ref{fig:comp_l2}, we empirically find that $\mathcal{L}_{2,LL}$ of the previous GAN inversion models have around 30 times larger value than $\mathcal{L}_{2,f\in\mathbb{F}_{h}}$, on average. 
This indicates that $\mathcal{L}_2$ predominantly focuses on the low-frequency information.

Recent GAN inversion model which targets high-fidelity inversion \cite{wang2022high}, uses $\mathcal{L}_{1}$ instead of $\mathcal{L}_{2}$.
Though $\mathcal{L}_{1}$ is known to be more robust than $\mathcal{L}_{2}$, it still struggles with frequency bias.
When we assume distributions of pixel-wise differences between $I_{1}$ and $I_{2}$ are i.i.d., and follow $ \mathcal{N} (\mu\thickapprox0, \sigma^2)$, $\mathcal{L}_{1}$ can be indicated as below:
\begin{align}
4 \log\mathbb{E}[\mathcal{L}_{1}(I_{1},I_{2})] + C = \sum_{f\in \mathbb{F}_{l}\cup\mathbb{F}_{h}} \lambda_{f}\log\mathbb{E}[\mathcal{L}_{1,f}(I_{1},I_{2})],
\end{align}
$\text{holds when}\, \lambda_{f}=1, \forall f\in\mathbb{F}_{l}\cup \mathbb{F}_{h} \text{ and } C \text{ is a constant.}$
Please refer to the appendix for the detailed proof.
Similar to $\mathcal{L}_{2}$, $\mathcal{L}_{1}$ yields same $\lambda_{f}$ for all $f$, which eventually leads to the frequency bias.

Consequently, we argue that
$\lambda_{f\in\mathbb{F}_{h}}$ should be higher than $\lambda_{LL}$ to avoid the low-frequency bias. We propose \textit{wavelet loss}, to focus on the high-frequency details.
The wavelet loss $\mathcal{L}_{wave}$ between $I_{1}$ and $I_{2}$ is defined as below:
\begin{equation}
\label{eq:wave}
\mathcal{L}_{p,wave}(I_{1},I_{2}) = \sum_{f\in\mathbb{F}_{h}}\mathcal{L}_{p,f}(I_{1},I_{2}),
\end{equation}
where $p \in \{1, 2\}$ for $\mathcal{L}_1$ and $\mathcal{L}_2$ loss.
$I_{1}$ and $I_{2}$ in equation \ref{eq:wave} only pass $f\in \mathbb{F}_{h}$, which discards the sub-bands with the frequency below $f_{nyq}/2$, where $f_{nyq}$ is the Nyquist frequency of the image.
However, we empirically find that a substantial amount of image details are placed below $f_{nyq}/2$ (See appendix for the qualitative image details in each sub-band).
In other words, $\mathcal{L}_{wave}$ should cover a broader range of frequency bands.
To this end, we improve $\mathcal{L}_{wave}$ combining with multi-level wavelet decomposition \cite{liu2018multi, yoo2019photorealistic}, which can subdivide the frequency ranges by iteratively applying four filters to the $LL$ sub-bands.
We improve $\mathcal{L}_{wave}$ with the $K$-level wavelet decomposition, named $\mathcal{L}^{K}_{wave}$:
\begin{equation}
\label{eq:wavek}
\mathcal{L}^{K}_{p, wave}(I_{1},I_{2}) = \sum_{i=1}^{K}\sum_{f\in \mathbb{F}_{h}}
\mathcal{L}_{p,f}((I_{1}*LL^{(i)}), (I_{2}*LL^{(i)})),
\end{equation}
where $p \in \{1, 2\}$ and $LL^{(i)}$ stands for passing $LL$ for $i$ times iteratively for multi-level wavelet decomposition.
$\mathcal{L}^{K}_{wave}$ can cover the image sub-bands with the frequency ranges between $f_{nyq}/2^{K}$ and $f_{nyq}$.

Since $\mathcal{L}^{K}_{wave}$ is specialized for preserving high-frequency details, we utilize it in the following two components of training:
First, we use it to compare the final inversion results with the ground truth, $\mathcal{L}^{K}_{2, wave}(X,\hat{X})$, to preserve high-frequency details at $\hat{X}$.
Second, we use $\mathcal{L}^{K}_{1, wave}$ for the direct high-frequency information transfer module, $ADA$.
The majority of high-frequency details in real-world images cannot be generated via prior knowledge of the generator \cite{alaluf2021hyperstyle}, which therefore should be transferred through additional modules, \ie, $ADA$ in WINE.
Consequently, mitigating the frequency bias on training $ADA$ is crucial.
To this end, we add the wavelet loss to train $ADA$:
\begin{equation}
\label{eq:lwave}
    \mathcal{L}_{wave,ADA}(\Delta, \hat{\Delta})=\mathcal{L}_{1}(\Delta, \hat{\Delta})+\lambda_{wave,ADA}\mathcal{L}^{K}_{1, wave}(\Delta, \hat{\Delta}).
\end{equation}
We set $\lambda_{wave,ADA}=0.1$ and $K=3$ for training.

The overall loss merges self-supervised alignment loss $\mathcal{L}_{wave,ADA}$ and reconstruction loss $\mathcal{L}_{image}$ between $X$ and the final image $\hat X$. 
Reconstruction loss consists of the weighted sum of $\mathcal{L}_2$, $\mathcal{L}_{2,wave}^K$, $\mathcal{L}_{id}$, and LPIPS \cite{zhang2018lpips}, with weights $\lambda_{\mathcal{L}_2}$, $\lambda_{wave}$, $\lambda_{id}$, and $\lambda_{LPIPS}$, respectively:
\begin{align}
\label{eq:trainloss}
\nonumber
\mathcal{L}_{total} = &\mathcal{L}_{wave, ADA}(\Delta, \hat{\Delta})\\
& + \mathcal{L}_{Image}(\lambda_{\mathcal{L}_{2}}, \lambda_{wave}, \lambda_{id}, \lambda_{LPIPS})(X,\hat{X}).
\end{align}

\begin{figure*}[ht]
\begin{center}
\includegraphics[width=1\textwidth]{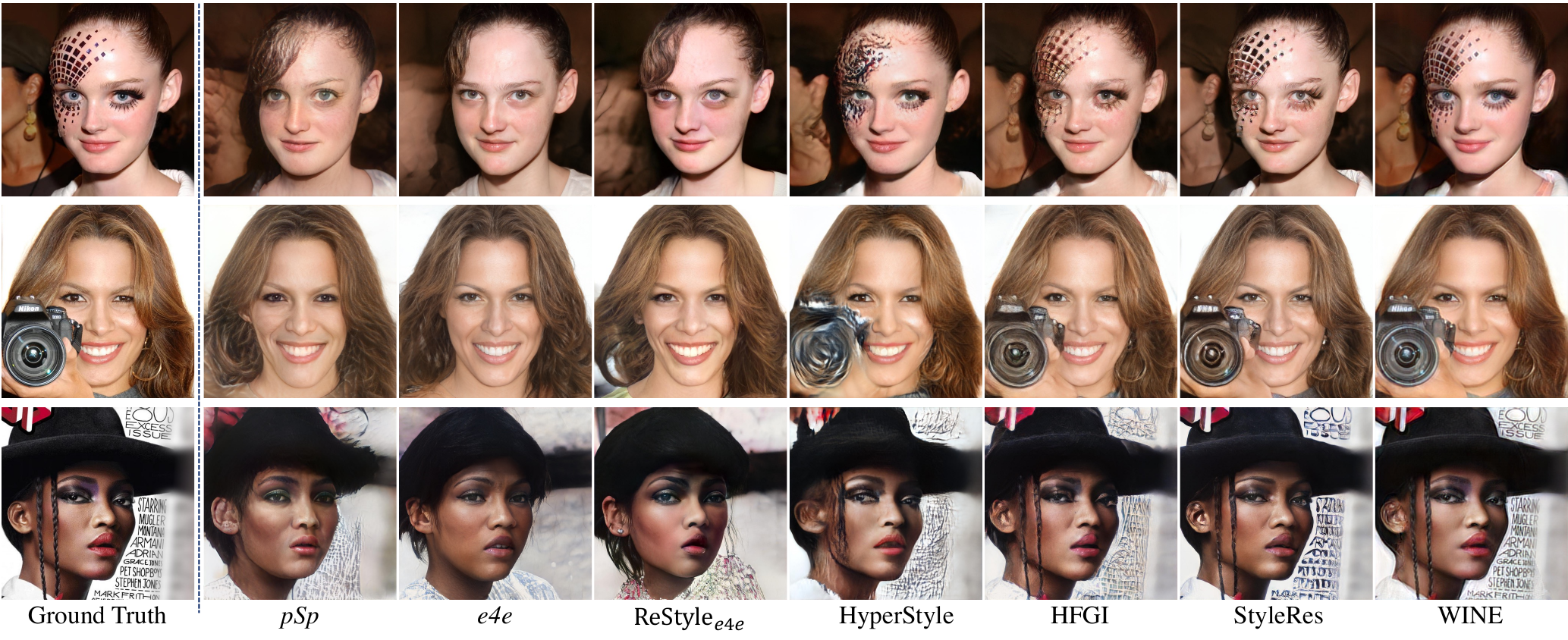}
\end{center}
\vspace{-0.4cm}
\caption{
\textbf{Qualitative Comparison between Inversion Results of Baselines.}
The baseline models including the state-of-the-art high-rate inversion models failed to preserve details, such as accessories and complex backgrounds.
In contrast, inverted images through WINE showed robust reconstruction of image-wise details, \eg, eyelashes, camera, and legible letters for each row.
}
\label{fig:inversion}
\vspace{-0.4cm}
\end{figure*}

\subsection{Wavelet Fusion}
\label{sec:method3}
\noindent To prevent the generator from relying on the low-rate latent, \ie, $w\in W+$, we should delicately transfer information from $\Tilde{\Delta}$ to the generator.
For instance, HFGI \cite{wang2022high} extracts \textit{scale} $g_{feat}$ and \textit{shift} $h_{feat}$ using $\Tilde{\Delta}$,
and fuses them with $F_{\ell_{f}}$, the original StyleGAN intermediate feature at layer $\ell_f$, and the latent at layer $\ell_{f}$, $w_{\ell_f}$ as follows:
\begin{equation}\label{eq:featurefusion}
F_{\ell_f+1} = g_{feat} \cdot \text{ModConv}(F_{\ell_f}, w_{\ell_f}) + h_{feat}.
\end{equation}
Though feature fusion is helpful for preserving the image-specific details, the majority of image details, \eg, exact boundaries, are still collapsed
(see qualitative evaluation for more details).
We attribute this to the low-resolution of feature fusion.
In the case of HFGI, the feature fusion is only applied to the resolution of 64, \ie, $\ell_f=7$.
According to the Shannon-Nyquist theorem, the image $I \in \mathbb{R}^{H\times W} (H=W=l)$, cannot store information with the frequency range higher than $f_{nyq} = \sqrt{H^2+W^2}=l\sqrt{2}$.
Consequently, the upper bound of information frequency by the feature fusion of HFGI is $f_{nyq}=64\sqrt{2}$, which is relatively lower than the image size, $1024$.
To solve this, we modify the feature fusion to be done on both 64 and 128 resolutions, i.e., $\ell_f$ = 7 and 9, respectively, which means $f_{nyq}$ is doubled.

However, a simple resolution increment cannot address the problem totally.
Since feature fusion goes through the pre-trained generator layers, the degradation of image-specific details is inevitable.
To this end, we propose a novel method, named \textit{wavelet fusion}.
Wavelet fusion directly handles the wavelet coefficients instead of the generator feature.

Using the hierarchical upsampling structure of SWAGAN which explicitly constructs the wavelet coefficients, wavelet fusion can transfer high-frequency knowledge without degradation.
Similar to feature fusion, our model obtains \textit{scale} $g_{wave}$ and \textit{shift} $h_{wave}$ using $\Tilde{\Delta}$, and fuse them with the wavelet coefficients obtained by the layer, tWavelets as:
\begin{equation}\label{eq:wavfusion}
\mathcal{W}_{\ell_w} = g_{wave} \cdot \text{tWavelets}(F_{\ell_w}) + h_{wave},
\end{equation}
where $\mathcal{W}_{\ell_w}$ are the wavelet coefficients of the $\ell_{w}$-th layer.

We empirically find that feature fusion is helpful for reconstructing the coarse shape, while wavelet fusion is for the fine details.
Consequently, we apply feature fusion for earlier layers ($\ell_{f}=\{7,9\}$) than wavelet fusion ($\ell_{w}=11$).
See Section \ref{sec:ablation} for ablations of the fusion layer selection.

\begin{figure*}[!t]
\begin{center}
\includegraphics[width=0.85\textwidth]{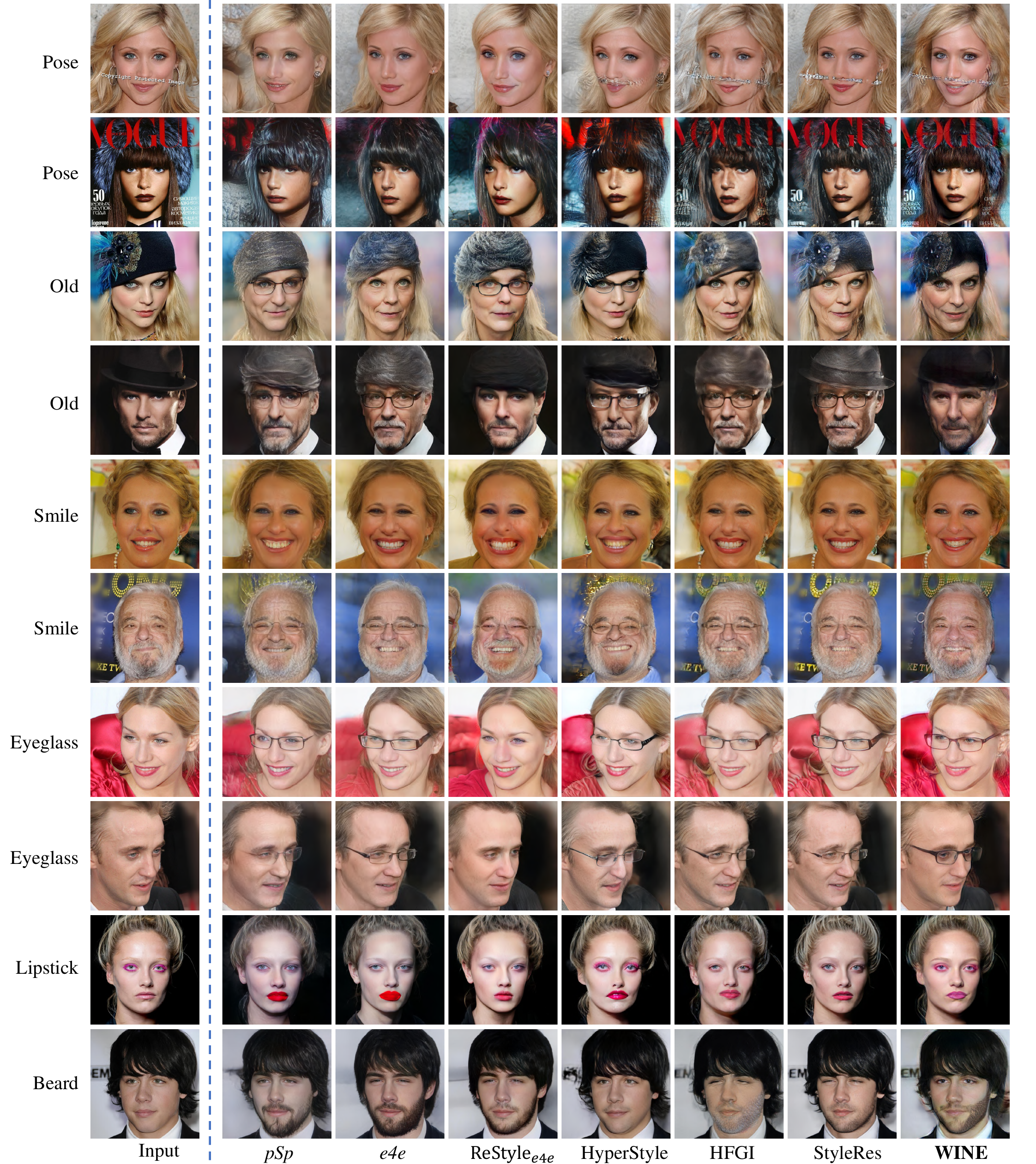}
\vspace{-0.3cm}
\end{center}
\caption{
\textbf{Qualitative Comparison between Editing Results of Baselines.}
We show edited images via InterFaceGAN (1-8th rows) and editing results via StyleCLIP (9-10th rows).
Both low- and high-rate inversion baselines suffered from preserving details, while our proposed method efficiently restored high-fidelity details with satisfactory editability with highly disentangled editing performance.
}
\label{fig:editing}
\vspace{-0.21cm}
\end{figure*}

\section{Experiments}
\label{experiments}

\noindent In this section, we compared the results of WINE with various GAN inversion methods.
We used the widely-used low-rate inversion models, \eg, $pSp$ \cite{richardson2021encoding},  $e4e$ \cite{tov2021designing}, and ReStyle \cite{alaluf2021restyle}, together with the state-of-the-art high-rate inversion models, \eg, HyperStyle \cite{alaluf2021hyperstyle}, HFGI \cite{wang2022high}, and StyleRes \cite{pehlivan2023styleres} as baselines.
First, we compared the qualitative results of inversion and editing scenarios. For editing, we used InterFaceGAN \cite{shen2020interfacegan} and StyleCLIP \cite{patashnik2021styleclip} to manipulate the latents.
Next, we compared the inversion performance quantitatively.
Finally, we analyzed the effectiveness of our wavelet loss and wavelet fusion through tactical ablation studies.
We first conducted experiments in the human face domain: we used Flickr-Faces-HQ (FFHQ) dataset \cite{karras2019style} for training and CelebA-HQ dataset \cite{Karras2018celeb} for evaluation, with all images generated to the high-resolution $1024^{2}$. We also provide experimental results on Animal-Faces-HQ (AFHQ) dataset \cite{choi2020starganv2}. We trained our own $e4e$ encoder, InterFaceGAN, and StyleCLIP for each attribute to exploit the latent space of SWAGAN \cite{gal2021swagan}.
Additional ablation experimental results of WINE with varying loss weights, fusion layers, and fusion methods are provided in the appendix.



\begin{figure}[t]
\begin{center}
\includegraphics[width=\columnwidth]{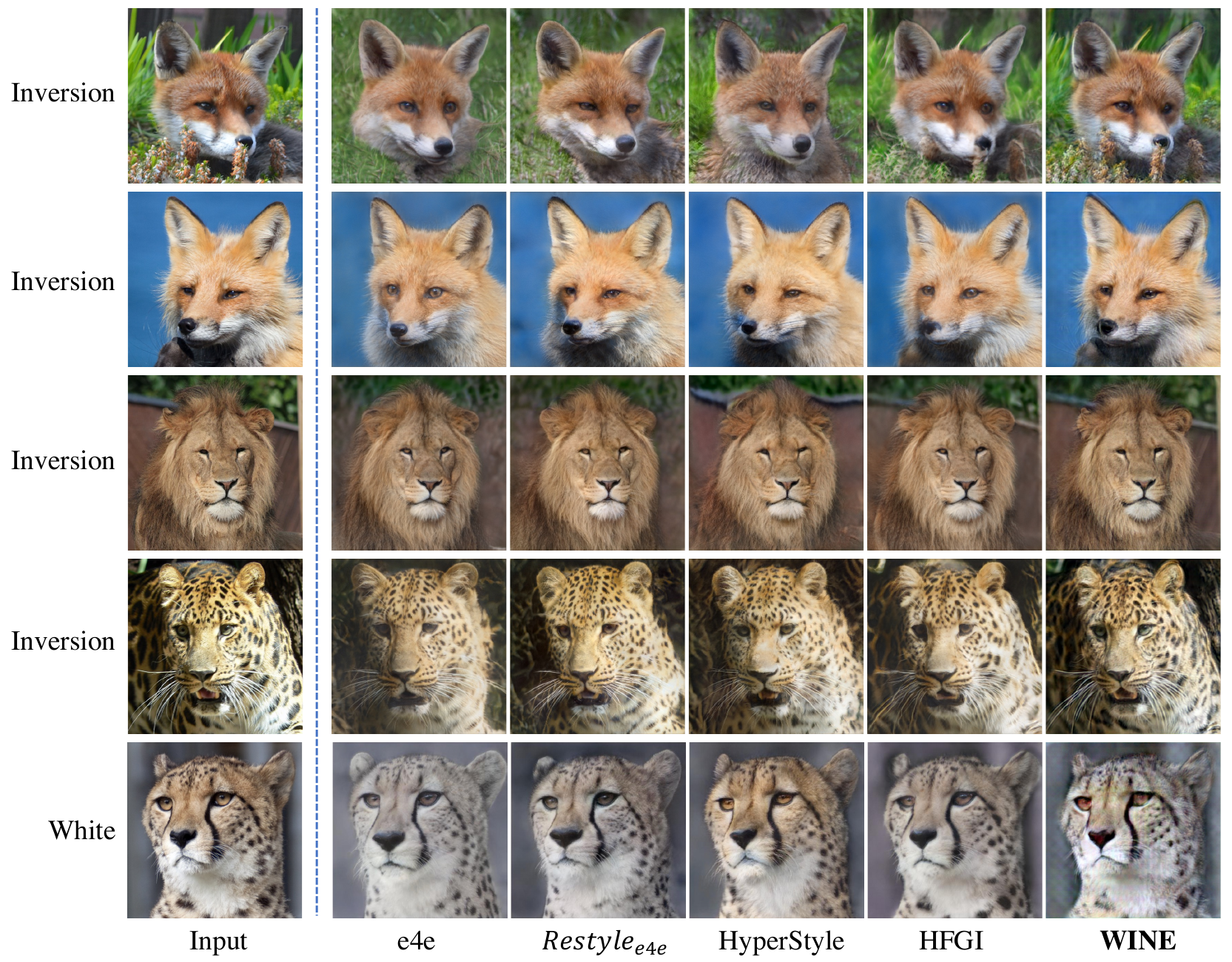}
\end{center}
\vspace{-0.3cm}
\caption{\textbf{Qualitative Comparison of GAN Inverted AFHQ Dataset Images.} We show the inverted images in the first four rows and manipulated images via StyleCLIP in the last row. Only WINE is capable of reconstructing and editing without losing high-frequency details and identity, even for non-human images.
}
\label{fig:appendix_editing_afhq}
\end{figure}

\subsection{Qualitative Evaluation}
\label{exp:qual}

\begin{table}[!t]
\small
\centering
\resizebox{\columnwidth}{!}{%
\begin{tabular}{l|ccccc}
\toprule
Method  & $\mathcal{L}_2$ $\downarrow$ & $\mathcal{L}_{1,wave}$ $\downarrow$ & LPIPS $\downarrow$ & SSIM $\uparrow$ & ID $\uparrow$ \\
\midrule
$pSp$ & 0.039 & 0.369 & 0.161 & 0.644 & 0.810\\
$e4e$ & 0.053 & 0.390 & 0.198 & 0.591 & 0.786    \\
$\text{ReStyle}$ & 0.049 & 0.389 & 0.190 &  0.638 & 0.783 \\
HS & 0.027 & 0.354 & 0.098 & 0.652 & 0.831 \\
HFGI & 0.023 & 0.351 & 0.091 & 0.661  & 0.864  \\
StyleRes & 0.018 & 0.301 & 0.076 & 0.708 & 0.894 \\
\textbf{WINE} & \textbf{0.011} & \textbf{0.230} & \textbf{0.051} & \textbf{0.753} & \textbf{0.906} \\
\bottomrule
\end{tabular}
}
\caption{\textbf{Quantitative Comparison between Inversion Results of Baselines.}
WINE consistently outperformed all baselines remarkably in various metrics related to either distortion or fidelity.
Quantitative results showed that our model learned the ground truth frequency distribution most accurately, without loss of identity while maintaining high perceptual quality.}
\vspace{-0.3cm}
\label{tab:inversion}
\end{table}

\noindent We first show the qualitative comparisons of inversions and editings in Figure \ref{fig:inversion} and Figure \ref{fig:editing}.  
We observed that our model produced more realistic inversion results than baselines, especially when images required more reconstructions of fine-grained details or complex backgrounds.
For instance, HyperStyle \cite{alaluf2021hyperstyle}, which refined weights of the generator per image, failed to reconstruct out-of-distribution objects, \eg, facial tattoo, and camera.
HFGI \cite{wang2022highcode} and StyleRes \cite{pehlivan2023styleres} could generate most of the lost details from the initial inversion, but restored details were close to artifacts, \eg, background letters.
In contrast, our method reconstructed details with minimum distortion consistently.

Figure \ref{fig:editing} shows editing results for six attributes, manipulated by InterFaceGAN \cite{shen2020interfacegan} and StyleCLIP \cite{patashnik2021styleclip}. 
Our model consistently showed the most robust results with high editability, while the baselines failed to either edit images or restore details.
For instance, in the case of InterFaceGAN, the baselines struggled to preserve details, \eg, the hats in the second to fourth rows.
In the case of StyleCLIP, the baselines similarly failed to preserve details, let alone address issues related to identity shifts.
Overall, both InterFaceGAN and StyleCLIP editing results showed that our model is the most capable of handling the trade-off between reconstruction quality and editability.

Figure \ref{fig:appendix_editing_afhq} illustrates a comparison between inverted and edited AFHQ images \cite{choi2020starganv2}, with baselines that offer pre-trained checkpoints or training codes. The animal images inverted by WINE (rows one to four) maintain intricate details such as whiskers, eyes, and backgrounds. Notably, the last row displays images edited using StyleCLIP, and WINE stands out as the sole method capable of reconstructing images without losing specific details.

\begin{figure}[t]
\begin{center}
\includegraphics[width=\columnwidth]{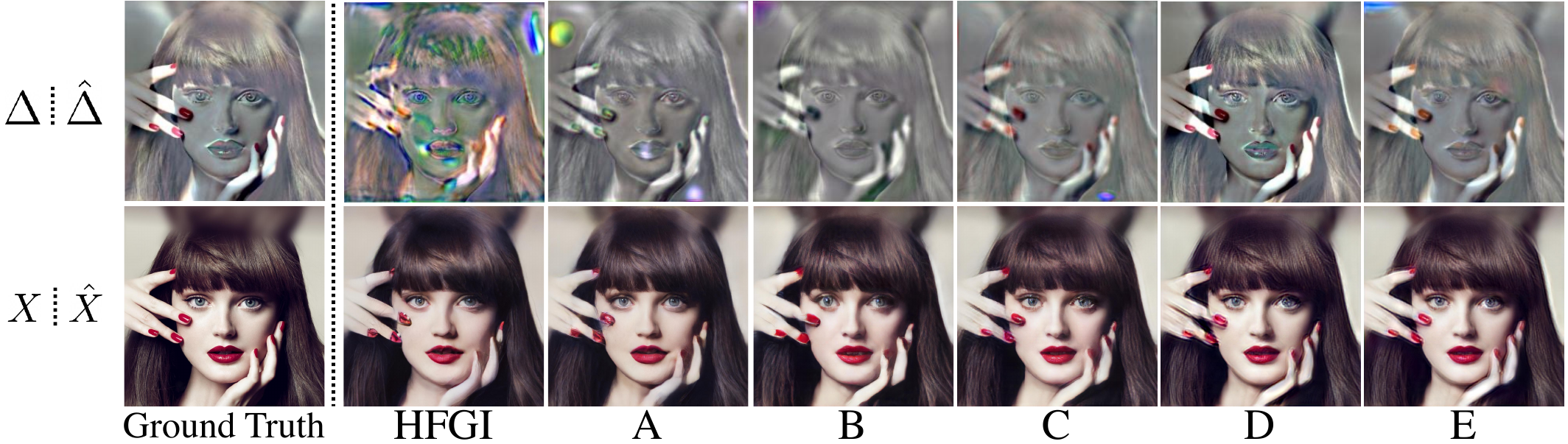}
\end{center}
\vspace{-0.3cm}
\caption{
\textbf{Qualitative Ablation of WINE.} We compared the performance of $ADA$ and the final inversion from the visualization of $\hat\Delta$ and $\hat X$, together with $\Delta$ and $X$.
While the existing state-of-the-art model introduced artifacts for computing $\hat\Delta$, our proposed methods reduced artifacts effectively and showed better inversion.
}
\label{fig:ablation}
\vspace{-0.5cm}
\end{figure}

\subsection{Quantitative Evaluation}
\noindent We evaluated inversion qualities with the existing baselines. Table \ref{tab:inversion} shows the quantitative results of each method. We used all images in the test split of CelebA-HQ and evaluated them with (i) the standard objectives including $\mathcal{L}_2$, LPIPS and ID similarity,
and (ii) the wavelet loss (equation \ref{eq:wavek}) to measure the spatial frequency distortion in the high-frequency sub-band.
Our model consistently outperformed every baseline by a large margin.
The results empirically prove that our wavelet loss and wavelet fusion in the frequency domain are capable of minimizing the distortion and thus improving the reconstruction quality. Please refer to the appendix for quantitative comparison of edited results.

\subsection{Ablation Study}
\label{sec:ablation}

\noindent In this section, we analyzed the effectiveness of each component of WINE, especially for our wavelet loss and wavelet fusion.
In Table \ref{tab:ablation}, we quantitatively compared the performance while changing each component.
We compared $\mathcal{L}_{1}$ through $\hat\Delta$ for the evaluation of $ADA$, and $\mathcal{L}_2$ and SSIM through $\hat X$ for the evaluation of the final inversion.

The \textbf{baseline} model without wavelets incorporated throughout the generator (StyleGAN2 \cite{karras2020analyzing}), and without our wavelet loss and wavelet fusion, corresponds to HFGI \cite{wang2022high}.
Firstly, configuration \textbf{A} only added the wavelet loss on HFGI.
We indeed found that the wavelet loss was helpful even when it was \textit{solely} used in the StyleGAN backbone.
In precise, \textsc{A} shows remarkably higher SSIM than HFGI.
Moreover, in Figure \ref{fig:ablation}, distortion in $\hat\Delta$ of \textsc{A} was decreased significantly, and details of $\hat X$ of \textsc{A}, \eg, fingers, were enhanced notably, compared to them of HFGI. For scenarios B to E, we adopted the generator with the hierarchical generation of wavelet coefficients (SWAGAN \cite{gal2021swagan}).
A simple modification of the generator from StyleGAN2 to SWAGAN (\textbf{B}) made marginal degradation in inversion.
We attribute this to the structure of existing inversion models, which are fitted to the StyleGAN.
In \textbf{C}, training with the wavelet loss achieved remarkably lower $\mathcal{L}_{1}(\Delta, \hat\Delta)$ (about 19\%) than training solely with $\mathcal{L}_{1}$.
Similar to A, this is compelling that training together with the wavelet loss achieved lower $\mathcal{L}_{1}$ than training solely with $\mathcal{L}_{1}$.
Moreover, as shown in Figure \ref{fig:ablation}, the wavelet loss was indeed helpful for preserving details, \eg, nail colors, compared to B.
With the residual transferred via wavelet fusion (\textbf{D}), though SSIM is improved, we can find ghosting artifacts in Figure \ref{fig:ablation}.
We assume that the absence of the wavelet loss refrains $ADA$ from finding the appropriate alignment, which leads to ghosting artifacts.
Finally, when \textbf{E} combined all the components together, \ie, WINE, apparent gain occurs on both $ADA$ and the final inversion.
Until \textsc{C}, though we elaborately calculated high-frequency features via wavelet loss, the model did not effectively transfer it to the generator.
The wavelet fusion at last enables the high-frequency information transfer to the generator, showing the synergistic effect with wavelet loss.

\begin{table}[t]
\small
\centering
\resizebox{\columnwidth}{!}{%
\begin{tabular}{l|ccc|ccc}
\toprule
\textbf{Config} & \makecell{Wavelet \\ Generator} & \makecell{Wavelet \\ Loss} & \makecell{Wavelet \\ Fusion} 

& $\mathcal{L}_{1}(\Delta, \hat{\Delta})\downarrow$ & $\mathcal{L}_{2}\downarrow$ & SSIM$\uparrow$ \\
\midrule
\normalsize HFGI &  & & - & 0.124 & 0.023 & 0.653 \\
\small\textsc{A} &  & \checkmark & - & 0.120 & 0.022 & 0.703 \\
\midrule
\small\textsc{B} & \checkmark  & & & 0.124 & 0.024 & 0.659 \\
\small\textsc{C} & \checkmark  & \checkmark & & 0.100 & 0.023 & 0.660 \\
\small\textsc{D} & \checkmark  & & \checkmark & 0.099 & 0.024 & 0.717 \\
\rowcolor[gray]{.9}
\textbf{\small\textsc{E} (WINE)} & \checkmark & \checkmark & \checkmark & \textbf{0.089} & \textbf{0.011} & \textbf{0.753} \\
\bottomrule
\end{tabular}
}
\caption{\textbf{Quantitative Ablation of WINE.}
We compared the performances of $ADA$ and the final inversion by adding each component we proposed.
Except for altering generators from StyleGAN to SWAGAN, our proposed methods, \ie, wavelet loss and fusion, showed remarkable gains in $ADA$ and the inversion, respectively.
}
\label{tab:ablation}
\vspace{-0.3cm}
\end{table}

\vspace{-0.1cm}

\section{Conclusion}
\label{conclusion}
\noindent Recent high-rate GAN inversion models focus on preserving image-wise details but still suffer from the low-frequency bias.
We pointed out that the existing methods are biased on low-frequency sub-band, in both structural and training aspects.
We proposed WINE, a novel GAN inversion and editing model, which explicitly handles the wavelet coefficients of the high-frequency sub-band via wavelet loss and wavelet fusion.
We demonstrated that WINE achieved the best performance among the state-of-the-art GAN inversions.
Moreover, we explored the effectiveness of each method through the ablation study.

\vspace{-0.1cm}
\section{Acknowledgments}
This work was supported by Institute of Information \& Communications Technology Planning \& Evaluation (IITP) grant funded by the Korea government (MSIT); (No.2022-0-00058, Development of sound-based photorealistic 3D face generation technology)
Also, This research was supported by Culture, Sports, and Tourism R\&D Program through the Korea Creative Content Agency grant funded by the Ministry of Culture, Sports and Tourism in 2024 (RS-2024-00441174), and MSIT (Ministry of Science and ICT), Korea, under the ITRC (Information Technology Research Center) support program (IITP-2024-RS-2023-00258649) supervised by the IITP (Institute for Information \& Communications Technology Planning \& Evaluation), and in part by the IITP grant funded by the Korea Government (MSIT) (Artificial Intelligence Innovation Hub) under Grant 2021-0-02068, and by the IITP grant funded by the Korea government (MSIT) (No.RS-2022-00155911, Artificial Intelligence Convergence Innovation Human Resources Development (Kyung Hee University)).

{\small
\bibliographystyle{ieee_fullname}
\bibliography{egbib}
}

\clearpage

\setcounter{page}{1}


\small

\section{Proofs for the Theorem}
\begin{strip}
\label{sec:app1}
In this section, we show the proof for the proposed equations in the main paper;

First, the following equation holds when
$\lambda_{f}=1$, $\forall f\in\displaystyle \mathbb{F}_{l}\cup\displaystyle \mathbb{F}_{h}$:
\begin{equation}
\label{eq:1}
\mathcal{L}_{2}(I_{1},I_{2}) = \sum_{f\in\displaystyle \mathbb{F}_{l}\cup\displaystyle \mathbb{F}_{h}}\lambda_{f}\mathcal{L}_{2,f}(I_{1},I_{2}).
\end{equation}
\begin{proof}
Let $I_{1}=(a_{ij})\in\displaystyle \mathbb{R}^{m\times n}$, and $I_{2}=(b_{ij})\in\displaystyle \mathbb{R}^{m\times n}$.

And for the simplicity of the notation, we define $c_{i,j}=a_{i,j}-b_{i,j}$.

Then,
$$\mathcal{L}_{2}(I_{1},I_{2})=\frac{1}{mn}\sum_{i\in\{0,1,\dots m\}}\sum_{j\in\{0,1,\dots n\}}(c_{ij})^2.$$
Let $m'=[\frac{m}{2}], n'=[\frac{n}{2}]$.
Then, we can denote $\mathcal{L}_{2,f\in\displaystyle \mathbb{F}_{l}\cup\displaystyle \mathbb{F}_{h}}$ as below;
\begin{align*}
& \mathcal{L}_{2,LL}(I_{1},I_{2})=\frac{1}{m'n'}\sum_{i\in\{0,1,\dots m'\}}\sum_{j\in\{0,1,\dots n'\}} (c_{2i+1,2j+1}+c_{2i+1,2j}+c_{2i,2j+1}+c_{2i,2j})^2, \\
& \mathcal{L}_{2,LH}(I_{1},I_{2})=\frac{1}{m'n'}\sum_{i\in\{0,1,\dots m'\}}\sum_{j\in\{0,1,\dots n'\}} (-c_{2i+1,2j+1}-c_{2i+1,2j}+c_{2i,2j+1}+c_{2i,2j})^2, \\
& \mathcal{L}_{2,HL}(I_{1},I_{2})=\frac{1}{m'n'}\sum_{i\in\{0,1,\dots m'\}}\sum_{j\in\{0,1,\dots n'\}} (-c_{2i+1,2j+1}+c_{2i+1,2j}-c_{2i,2j+1}+c_{2i,2j})^2, \\
& \mathcal{L}_{2,HH}(I_{1},I_{2})=\frac{1}{m'n'}\sum_{i\in\{0,1,\dots m'\}}\sum_{j\in\{0,1,\dots n'\}} (c_{2i+1,2j+1}-c_{2i+1,2j}-c_{2i,2j+1}+c_{2i,2j})^2.
\end{align*}

We can rewrite $\mathcal{L}_{2}(I_{1},I_{2})$ as;
$$\mathcal{L}_{2}(I_{1},I_{2})=\frac{4}{m'n'}\sum_{i\in\{0,1,\dots m'\}}\sum_{j\in\{0,1,\dots n'\}} c_{2i+1,2j+1}^2+c_{2i+1,2j}^2+c_{2i,2j+1}^2+c_{2i,2j}^2.$$
We use the following identical equation, which holds for $\forall x,y,z,w\in\displaystyle \mathbb{R}$; $$(x+y+z+w)^2+(-x-y+z+w)^2+(-x+y-z+w)^2+(x-y-z+w)^2 = 4(x^2+y^2+z^2+w^2).$$
We can obtain the following;

$$\sum_{f\in\displaystyle \mathbb{F}_{l}\cup\displaystyle \mathbb{F}_{h}}\mathcal{L}_{2,f}(I_{1},I_{2}) = \frac{1}{m'n'}\sum_{i\in\{0,1,\dots m'\}}\sum_{j\in\{0,1,\dots n'\}}4(c_{2i+1,2j+1}^2+c_{2i+1,2j}^2+c_{2i,2j+1}^2+c_{2i,2j}^2).$$
$$\therefore \mathcal{L}_{2}(I_{1},I_{2}) = \sum_{f\in\displaystyle \mathbb{F}_{l}\cup\displaystyle \mathbb{F}_{h}}1\cdot \mathcal{L}_{2,f}(I_{1},I_{2}).$$
\end{proof}


Second, When the distributions of pixel-wise differences between $I_{1}$ and $I_{2}$ are i.i.d., and follow $\displaystyle \mathcal{N} (\mu, \sigma^2)$ with $\mu\thickapprox0$, the following equation holds when 
$\lambda_{f}=1$, $\forall f\in\displaystyle \mathbb{F}_{l}\cup\displaystyle \mathbb{F}_{h}$:
\begin{equation}
\displaystyle 4\log\displaystyle  \mathbb{E}[\mathcal{L}_{1}(I_{1},I_{2})] + C = \sum_{f\in\displaystyle \mathbb{F}_{l}\cup\displaystyle \mathbb{F}_{h}}\displaystyle  \displaystyle \lambda_{f}\log\mathbb{E}[\mathcal{L}_{1,f}(I_{1},I_{2})],
\end{equation}
where $C$ is a constant.
\begin{proof}
Similar with proving equation \ref{eq:1}, we can derive followings;
\begin{align*}
& \mathcal{L}_{1}(I_{1},I_{2})=\frac{4}{m'n'}\sum_{i\in\{0,1,\dots m'\}}\sum_{j\in\{0,1,\dots n'\}}
|c_{2i+1,2j+1}|+|c_{2i+1,2j}|+|c_{2i,2j+1}|+|c_{2i,2j}|, \\
& \mathcal{L}_{1,LL}(I_{1},I_{2})=\frac{1}{m'n'}\sum_{i\in\{0,1,\dots m'\}}\sum_{j\in\{0,1,\dots n'\}} |c_{2i+1,2j+1}+c_{2i+1,2j}+c_{2i,2j+1}+c_{2i,2j}|, \\
& \mathcal{L}_{1,LH}(I_{1},I_{2})=\frac{1}{m'n'}\sum_{i\in\{0,1,\dots m'\}}\sum_{j\in\{0,1,\dots n'\}} |-c_{2i+1,2j+1}-c_{2i+1,2j}+c_{2i,2j+1}+c_{2i,2j}|, \\
& \mathcal{L}_{1,HL}(I_{1},I_{2})=\frac{1}{m'n'}\sum_{i\in\{0,1,\dots m'\}}\sum_{j\in\{0,1,\dots n'\}} |-c_{2i+1,2j+1}+c_{2i+1,2j}-c_{2i,2j+1}+c_{2i,2j}|, \\
& \mathcal{L}_{1,HH}(I_{1},I_{2})=\frac{1}{m'n'}\sum_{i\in\{0,1,\dots m'\}}\sum_{j\in\{0,1,\dots n'\}} |c_{2i+1,2j+1}-c_{2i+1,2j}-c_{2i,2j+1}+c_{2i,2j}|.
\end{align*}
Using $c_{i,j} \sim \displaystyle \mathcal{N} (\mu, \sigma^2)$, we can obtain followings:
$$(c_{2i+1,2j+1}+c_{2i+1,2j}+c_{2i,2j+1}+c_{2i,2j}) \sim \displaystyle \mathcal{N} (4\mu, 4\sigma^2),$$
$$(-c_{2i+1,2j+1}-c_{2i+1,2j}+c_{2i,2j+1}+c_{2i,2j}) \sim \displaystyle \mathcal{N} (0, 4\sigma^2),$$
$$(-c_{2i+1,2j+1}+c_{2i+1,2j}-c_{2i,2j+1}+c_{2i,2j}) \sim \displaystyle \mathcal{N} (0, 4\sigma^2),$$
$$(c_{2i+1,2j+1}-c_{2i+1,2j}-c_{2i,2j+1}+c_{2i,2j}) \sim \displaystyle \mathcal{N} (0, 4\sigma^2).$$

According to the properties of half-normal distribution,
for $p \sim \displaystyle \mathcal{N} (\mu, \sigma^2),$
$$\mathbb{E}[|p|]=\sigma\sqrt{\frac{2}{\pi}}e^{-\frac{\mu^2}{2\sigma^2}}+\mu\cdot\text{erf}(\frac{\mu}{\sqrt{(2\sigma^2)}}), \text{ where } \text{erf}(x)=\int_0^xe^{-t^2}dt.$$
Consequently,
\begin{align*}
& \mathbb{E}[|c_{2i+1,2j+1}|+|c_{2i+1,2j}|+|c_{2i,2j+1}|+|c_{2i,2j}|]=
4\sigma\sqrt{\frac{2}{\pi}}e^{-\frac{\mu^2}{2\sigma^2}}+4\mu\cdot\text{erf}(\frac{\mu}{\sqrt{(2\sigma^2)}}), \\
& \mathbb{E}[|c_{2i+1,2j+1}+c_{2i+1,2j}+c_{2i,2j+1}+c_{2i,2j}|]=
2\sigma\sqrt{\frac{2}{\pi}}e^{-4\cdot\frac{\mu^2}{2\sigma^2}}+4\mu\cdot\text{erf}(\frac{2\mu}{\sqrt{(2\sigma^2)}}), \\
& \mathbb{E}[|-c_{2i+1,2j+1}-c_{2i+1,2j}+c_{2i,2j+1}+c_{2i,2j}|]=
2\sigma\sqrt{\frac{2}{\pi}}, \\
& \mathbb{E}[|-c_{2i+1,2j+1}+c_{2i+1,2j}-c_{2i,2j+1}+c_{2i,2j}|]=
2\sigma\sqrt{\frac{2}{\pi}}, \\
& \mathbb{E}[|c_{2i+1,2j+1}-c_{2i+1,2j}-c_{2i,2j+1}+c_{2i,2j}|]=
2\sigma\sqrt{\frac{2}{\pi}}.
\end{align*}

Using the condition that $c_{i,j}$s are \textit{i.i.d.},
\begin{align*}
& \mathbb{E}[\mathcal{L}_{1}(I_{1},I_{2})] = \frac{4}{m'n'}\cdot m'n' \cdot (4\sigma\sqrt{\frac{2}{\pi}}e^{-\frac{\mu^2}{2\sigma^2}}+4\mu\cdot\text{erf}(\frac{2\mu}{\sqrt{2\sigma^2}})) \\
& = 16\sigma\sqrt{\frac{2}{\pi}}e^{-\frac{\mu^2}{2\sigma^2}}+16\mu\cdot\text{erf}(\frac{2\mu}{\sqrt{2\sigma^2}}), \\
& \mathbb{E}[\mathcal{L}_{1,LL}(I_{1},I_{2})] = \frac{1}{m'n'}\cdot m'n' \cdot (2\sigma\sqrt{\frac{2}{\pi}}e^{-4\cdot\frac{\mu^2}{2\sigma^2}}+4\mu\cdot\text{erf}(\frac{2\mu}{\sqrt{(2\sigma^2)}})) \\
& = 2\sigma\sqrt{\frac{2}{\pi}}e^{-4\cdot\frac{\mu^2}{2\sigma^2}}+4\mu\cdot\text{erf}(\frac{\mu}{\sqrt{(2\sigma^2)}}), \\
& \mathbb{E}[\mathcal{L}_{1,LH}(I_{1},I_{2})] = \mathbb{E}[\mathcal{L}_{1,HL}(I_{1},I_{2})] = \mathbb{E}[\mathcal{L}_{1,HH}(I_{1},I_{2})] = \frac{1}{m'n'}\cdot m'n' \cdot 2\sigma\sqrt{\frac{2}{\pi}} = 2\sigma\sqrt{\frac{2}{\pi}}. \\
\end{align*}

Since $\mu\thickapprox0$, 
$\mu\cdot\text{erf}(\frac{\mu}{\sqrt{2\sigma^2}})\thickapprox0$.
Consequently,
\begin{align*}
& \displaystyle \log\mathbb{E}[\mathcal{L}_{1}(I_{1},I_{2})] = \displaystyle \log16 + \displaystyle \log(\sigma\sqrt{\frac{2}{\pi}})-\frac{\mu^2}{2\sigma^2}, \\
& \displaystyle \log\mathbb{E}[\mathcal{L}_{1,LL}(I_{1},I_{2})] = \displaystyle \log2 + \displaystyle \log(\sigma\sqrt{\frac{2}{\pi}}) -4\cdot \frac{\mu^2}{2\sigma^2}, \\
& \displaystyle \log\mathbb{E}[\mathcal{L}_{1,LH}(I_{1},I_{2})] = \displaystyle \log\mathbb{E}[\mathcal{L}_{1,HL}(I_{1},I_{2})] = \displaystyle \log\mathbb{E}[\mathcal{L}_{1,HH}(I_{1},I_{2})] = 
\displaystyle \log2 + \displaystyle \log(\sigma\sqrt{\frac{2}{\pi}}).
\end{align*}
$$\displaystyle \log\mathbb{E}[\mathcal{L}_{1}(I_{1},I_{2})] =
\sum_{f\in\displaystyle \mathbb{F}_{l}\cup\displaystyle  \mathbb{F}_{h}}\displaystyle \frac{1}{4}\log\mathbb{E}[\mathcal{L}_{1,f}(I_{1},I_{2})] + C'.$$
$$\therefore \displaystyle 4\log\mathbb{E}[\mathcal{L}_{1}(I_{1},I_{2})] =
\sum_{f\in\displaystyle \mathbb{F}_{l}\cup\displaystyle  \mathbb{F}_{h}}\displaystyle \log\mathbb{E}[\mathcal{L}_{1,f}(I_{1},I_{2})] + C.$$

\end{proof}


\end{strip}
\begin{figure}[th]
\begin{center}
\includegraphics[width=0.7\columnwidth]{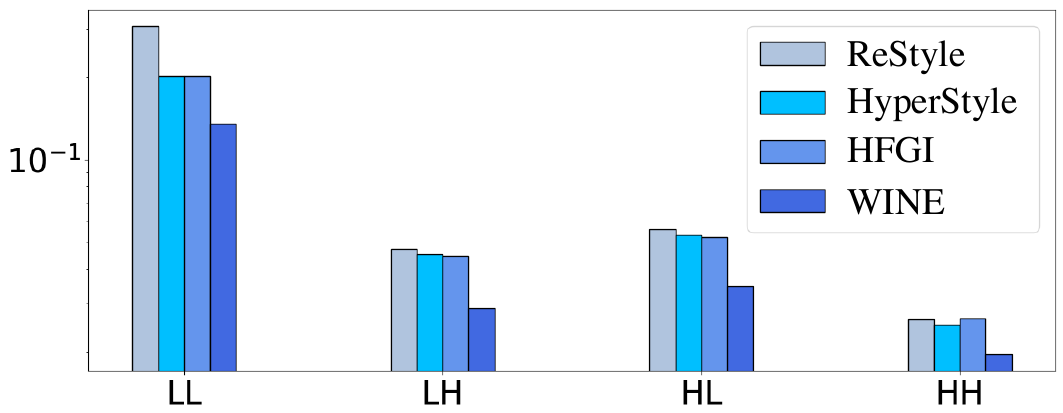}
\end{center}
\caption{
\textbf{Comparison of $\mathcal{L}_{1}$ of the wavelet coefficients.}
We plot the average $\mathcal{L}_{1}$ of each wavelet coefficient between CelebA-HQ test images and corresponding inverted images by various state-of-the-art inversion models. Due to the significant gap between $\mathcal{L}_{1,LL}$ and the rest (about 20 times in linear scale), we display the losses with the logarithmic scale for better visualization.
In contrast to other high-rate baseline inversion methods, \eg, HyperStyle and HFGI, WINE notably reduces distortion on high-frequency sub-bands.
}
\label{fig:l1_app}
\end{figure}
\begin{figure}[t!]
\begin{center}
\includegraphics[width=\columnwidth]{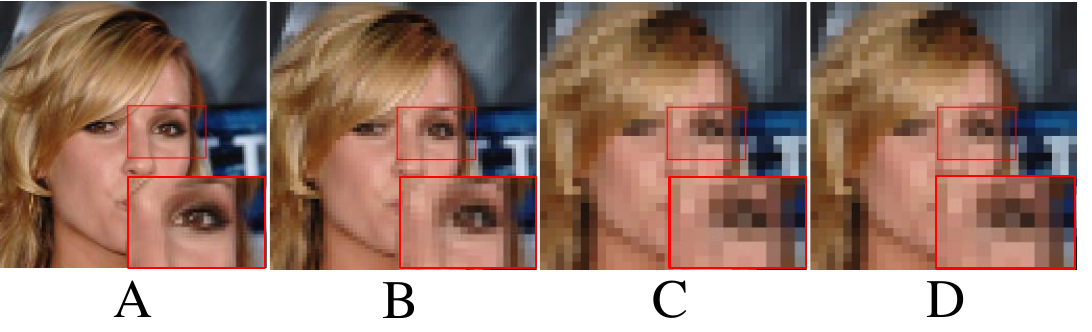}
\end{center}
\caption{
\textbf{Inverse wavelet transform results by omitting various wavelet sub-bands.
}
To check the qualitative image details in each sub-band, we remove the wavelet coefficients between $f_{nyq}/2$ $\sim$ $f_{nyq}$ (Config A), $f_{nyq}/2^2$ $\sim$ $f_{nyq}/2$ (Config B), and $f_{nyq}/2^3$ $\sim$ $f_{nyq}/2^2$ (Config C).
From A to B, severe degradation of visible image details does not occur.
However, for B to C or C to D, the majority of image details are degraded.
}
\label{fig:wavelet_omit}
\end{figure}

Similar to $\mathcal{L}_{2}$, $\mathcal{L}_{1}$ seems a fair loss without the frequency bias, which reflects $\mathcal{L}_{1,f\in\displaystyle \mathbb{F}_{l}\cup\displaystyle\mathbb{F}_{h}}$ with same weights.
However, as shown in Figure \ref{fig:l1_app}, we empirically find that $\mathcal{L}_{1,LL}$ is around 20 times larger than $\mathcal{L}_{1,f\in\displaystyle \mathbb{F}_{h}}$ in case of HyperStyle and HFGI.
This leads to the biased training, which results in an apparent decrease of $\mathcal{L}_{1,LL}$, but almost no gain, or even increment of $\mathcal{L}_{1,f\in\displaystyle \mathbb{F}_{h}}$, compared to state-of-the-art low-rate inversion method, \ie, ReStyle.
Consequently, we argue that $\mathcal{L}_{1}$ contains the low-frequency bias, and needs the wavelet loss to avoid it.

\subsection{Information in Sub-band of Images}
\label{app:subband}

In Section 3.2, we designed the multi-level wavelet loss to cover broader frequency ranges than $f_{nyq}/2$ $\sim$ $f_{nyq}$.
In Figure \ref{fig:wavelet_omit}, we show the results of the inverse wavelet transform by omitting the wavelet coefficients between $f_{nyq}/2$ $\sim$ $f_{nyq}$ (Config A), $f_{nyq}/2^2$ $\sim$ $f_{nyq}/2$ (Config B), and $f_{nyq}/2^3$ $\sim$ $f_{nyq}/2^2$ (Config C).
Though A removes the highest frequency sub-bands, \ie, $f_{nyq}/2$ $\sim$ $f_{nyq}$, among all configs, we cannot find visible degradation of image details.
In other words, information in the sub-band $f_{nyq}/2$ $\sim$ $f_{nyq}$ is mostly higher than the visible image details.
Since the firstly proposed wavelet loss (Equation 3 of the main paper) only covers the sub-band $f_{nyq}/2$ $\sim$ $f_{nyq}$, we should extend the range of sub-bands to effectively preserve the visible details.
Consequently, we propose a $K$-level wavelet loss, which enables covering the sub-band $f_{nyq}/2^{K}$ $\sim$ $f_{nyq}$.

\begin{center}
\begin{figure*}[!t]
\begin{center}
\includegraphics[width=1\textwidth]{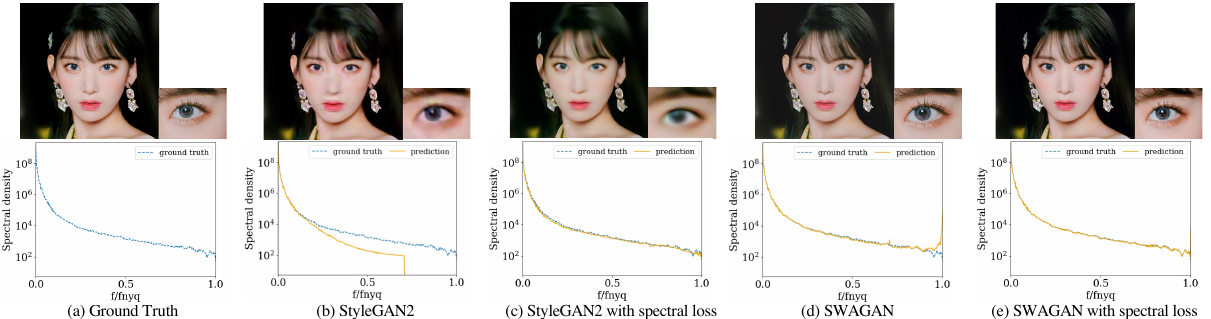}
\end{center}
\caption{
\textbf{Regression (top row) and spectral density plot (bottom row) of ground truth image and generated images trained with/without additional spectral loss.} Here, we used the spectral loss introduced in \cite{schwarz2021freqbias}. For both StyleGAN2 and SWAGAN generators, the additional spectral loss induced artifacts to coercively match the frequency distribution. We recommend you zoom in to carefully observe the reconstructed details.
}
\label{fig:freqbias}
\end{figure*}
\end{center}
\section{Generator Training with Spectral Loss}
\label{app:specloss}

Previous works \cite{durall2020upconv, jiang2021focal, schwarz2021freqbias} propose an objective function to precisely learn the frequency distribution of the training data, which we comprehensively named as \emph{spectral loss}. \cite{jiang2021focal} designed a spectral loss function that measures the distance between fake and real images in the frequency domain that captures both amplitude and phase information. \cite{durall2020upconv} proposed a spectral loss that measures the binary cross entropy between the azimuthal integration over the power spectrum of fake and real images. \cite{schwarz2021freqbias} used a simple $\mathcal{L}_2$ loss between the logarithm of the azimuthal average over power spectrum in normalized polar coordinates, \ie reduced spectrum, of fake and real images. We adopted the spectral loss term of \cite{schwarz2021freqbias} for our experiment :
\begin{equation} \label{eq:specloss}
\mathcal{L}_S = \frac{1}{H/\sqrt{2}} \sum_{k=0}^{H/\sqrt{2}-1} \|\log{(\tilde{S}(G(z)))}[k] - \log{(\tilde{S}(\mathbf{I}))}[k] \|_2^2, 
\end{equation}
where $\tilde{S}$ is the reduced spectrum, $G(z)$ is the generated image, and $\mathbf{I}$ is the ground truth real image. 

Here, we conducted a single-image reconstruction task, which is widely done \cite{gal2021swagan, schwarz2021freqbias} to investigate the effectiveness of explicit frequency matching in refining high-fidelity details. For StyleGAN2 \cite{karras2020analyzing} and SWAGAN \cite{gal2021swagan} generator, we used the latent optimization \cite{karras2019style} method to reconstruct a single image, each with and without the spectral loss. All images are generated to resolution $ 512 \times 512$, with the weight of spectrum loss $\times$0.1 of the original $\mathcal{L}_2$ loss. 

Figure \ref{fig:freqbias} shows the reconstructed images and spectral density plots for each case. As seen in Figure \ref{fig:freqbias}(a), the spectrum of a natural image follows an exponential decay. Using $\mathcal{L}_2$ singularly made both StyleGAN2 and SWAGAN generators overfit to the mostly existing low-frequency distribution.
(b) StyleGAN2 struggled to learn the high-fidelity details, creating an unrealistic image.
(d) SWAGAN was capable of fitting most of the high-frequency parts, except created some excessive high-frequency noise due to checkerboard patterns.
Though utilizing the spectral loss for both generators (c,e) exquisitely matched all frequency distributions, qualitative results were degraded.
Matching the frequency induced unwanted artifacts to the images, and caused the degradation.
Due to the absence of the spatial information, the loss based on the spectral density inherently cannot reconstruct high-frequency details.
Comparably, our wavelet loss minimizes the $\mathcal{L}_1$ distance of high-frequency bands in the spatial frequency domain, restoring meaningful high-fidelity features.


\section{Experimental Details}
\subsection{Training Details}
In our experiments, we implement our experiments based on the pytorch-version code \footnote{https://github.com/rosinality/stylegan2-pytorch} for SWAGAN \cite{gal2021swagan}.
We converted the weights of pre-trained SWAGAN generator checkpoint from the official TensorFlow code\footnote{https://github.com/rinongal/swagan} to pytorch version.
We trained our model on a single GPU and took only 6 hours for the validation loss to saturate, whereas other StyleGAN2-based baselines required more than 2 days of training time.

Here, we explain the details of our reconstruction loss terms: $\mathcal{L}_2$, $\mathcal{L}_{id}$, and $\mathcal{L}_{LPIPS}$. We leverage $\mathcal{L}_2$, as it is most effective in keeping the generated image similar to the original image pixel-wise.
$\mathcal{L}_{id}$ is an identity loss defined as:
\begin{equation}
    \mathcal{L}_{id} = 1-<R(G_0(w)), R(\mathbf{I}) > ,
\end{equation}
where $R$ is the pre-trained ArcFace \cite{deng2021arcface} model, and $\mathbf{I}$ is the ground truth image. $\mathcal{L}_{id}$ minimizes the cosine distance between two face images to preserve the identity. LPIPS \cite{zhang2018lpips} enhances the perceptual quality of the image by minimizing the distance on the feature space of ImageNet \cite{deng2009imagenet} pre-trained network.
For training, we used weights $\lambda_{\mathcal{L}_2}$=1,  $\lambda_{id}$=0.1, $\lambda_{LPIPS}$=0.8, respectively, which follows the widely adopted experimental setups in previous GAN inversion methods. 

\begin{table}[t]

\centering\footnotesize
\resizebox{\columnwidth}{!}{
\small
\begin{tabular}{c|ccccc}
\toprule
$\lambda_{wave,ADA}$ & 0 & 0.01 & 0.05 & \textbf{0.1} & 0.5\\
\midrule

$\mathcal{L}_{2}$ $\downarrow$ & 0.024 & 0.017  & 0.15 & \textbf{0.011} & 0.021\\ 
$\mathcal{L}_1,{wave}$ $\downarrow$ & 0.274 & 0.249 & 0.243 &\textbf{ 0.230} & 0.248\\ 
SSIM $\uparrow$ & 0.717 & 0.724 & 0.730 & \textbf{0.753}& 0.719 \\
\bottomrule
\end{tabular}
}

\caption{Quantitative comparison with various wavelet loss ratio.}
\label{tab:wavelet_ratio}
\vspace{-0.5cm}
\end{table}

In Table \ref{tab:wavelet_ratio}, we show the effect of our proposed wavelet loss via adjusting the weight $\lambda_{wave,ADA}$ respective to the weight $\lambda_{\mathcal{L}_1}$ in Eq. 6 of the main paper. Reminder that the ADA loss aims to minimize the discrepancy in residual wavelet features. Increasing the weight $\lambda_{wave,ADA}$ up to $0.1$ shows that incorporating the wavelet loss effectively enhances the reconstruction of image-wise details, particularly in high-frequency regions. However, exceeding a weight of 0.1 leads to a decline in performance, as most image information resides in the low-frequency sub-band. In general, we applied balanced weights that effectively reconstruct high-frequency sub-bands without compromising the generation of low-frequency sub-bands.

\subsection{Dataset Description}
In this section, we describe the datasets used for experiments in the main paper.

\textbf{Flickr-Faces-HQ (FFHQ) dataset.} Our model and all baselines are trained with FFHQ \cite{karras2019style}, a well-aligned human face dataset with 70,000 images of resolution 1024 $\times$ 1024. FFHQ dataset is widely used for training various unconditional generators \cite{karras2019style, karras2020analyzing, karras2021alias}, and GAN inversion models \cite{richardson2021encoding, tov2021designing, alaluf2021restyle, alaluf2021hyperstyle, moon2022interestyle, wang2022high}. All of the baselines we used in the paper use the FFHQ dataset for training, which enables a fair comparison.

\textbf{CelebA-HQ dataset.}
CelebA-HQ dataset contains 30,000 human facial images of resolution 1024 $\times$ 1024, together with the segmentation masks.
Among 30,000 images, around 2,800 images are denoted as the test dataset.
We use the official split for the test dataset, and evaluate every baseline and our model with all images in the test dataset.

\textbf{Animal-Faces-HQ (AFHQ) dataset.}
AFHQ \cite{choi2020starganv2} dataset contains 15,000 high-quality images of cats, dogs, and wildlife animals at 512 × 512 resolution. We used 5000 images of wild animals for training WINE, and the test-set for evaluation.

\subsection{Baseline Descriptions}
\label{app:bmd}

In this section, we describe the existing GAN inversion baselines, which we used for comparison in Section 4. We exclude the model which needs image-wise optimization, such as Image2StyleGAN \cite{abdal2019image2stylegan} or Pivotal Tuning \cite{roich2021pivotal}.

$\textbf{pSp}$ pixel2Style2pixel ($pSp$) adopts pyramid \cite{lin2017feature} network for the encoder-based GAN inversion.
$pSp$ achieves the state-of-the-art performance among encoder-based inversion models at the time.
Moreover, $pSp$ shows the various adaptation of the encoder model to the various tasks using StyleGAN, such as image inpainting, face frontalization, or super-resolution.

\textbf{e4e} encoder4editing ($e4e$) proposes the existence of the trade-off between distortion and the perception-editability of the image inversion.
In the other words, $e4e$ proposes that the existing GAN inversion models which focus on lowering distortion, sacrifice the perceptual quality of inverted images, and the robustness on the editing scenario.
$e4e$ suggests that maintaining the latent close to the original StyleGAN latent space, \ie, $W$, enables the inverted image to have high perceptual quality and editability.
To this end, $e4e$ proposes additional training loss terms to keep the latent close to $W$ space.
Though distortion of $pSp$ is lower than $e4e$, $e4e$ shows apparently higher perceptual quality and editabilty than $pSp$.

\textbf{ReStyle} ReStyle suggests that a single feed-forward operation of existing encoder-based GAN inversion models, \ie, $pSp$ and $e4e$, is not enough to utilize every detail in the image.
To overcome this, ReStyle proposes an iterative refinement scheme, which infers the latent with feed-forward-based iterative calculation.
The lowest distortion that Restyle achieves among encoder-based GAN inversion models shows the effectiveness of the iterative refinement scheme.
Moreover, the iterative refinement scheme can be adapted to both $pSp$ and $e4e$, which enables constructing models that have strengths in lowering distortion, or high perceptual quality-editability, respectively.
To the best of our knowledge, $\text{ReStyle}_{pSp}$ achieves the lowest distortion among encoder-based models which do not use generator-tuning method\footnote{IntereStyle \cite{moon2022interestyle} achieves lower distortion on the \textit{interest region} than $\text{ReStyle}_{pSp}$, but not for the whole image region.}.
Since we utilize baselines that achieve lower distortion than $\text{ReStyle}_{pSp}$, \ie, HyperStyle and HFGI, we only use $\text{ReStyle}_{e4e}$ to evaluate its high editability.

\textbf{HyperStyle} To make a further improvement from ReStyle, Pivotal Tuning \cite{roich2021pivotal} uses the input-wise generator tuning.
However, this is extremely time-consuming, and inconvenient in that it requires separate generators per every input image.
To overcome this, HyperStyle adopts HyperNetwork \cite{ha2016hypernetworks}, which enables tuning the convolutional weights of pre-trained StyleGAN only with the feed-forward calculation.
Starting from the latent obtained by $e4e$, HyperStyle iteratively refines the generator to reconstruct the original image with the fixed latent.
HyperStyle achieves the lowest distortion among encoder-based GAN inversion models at the time.

\textbf{HFGI} HFGI points out the limitation of the low-rate inversion methods and argues that encoders should adopt larger dimensions of tensors to transfer high-fidelity image-wise details.
To achieve this, HFGI adapts feature fusion, which enables mixing the original StyleGAN feature with the feature obtained by the image-wise details.

\textbf{StyleRes} StyleRes handles the trade-off between the reconstruction and editing quality of real images. In order to obtain high-quality editing in high-rate latent spaces, StyleRes learns residual features in higher latent codes and how to transform these residual features to adapt to latent code manipulations. StyleRes achieves the lowest distortion among every GAN inversion method, except our model.

\clearpage

\subsection{Quantitative Comparison on Editability}

\begin{table}[t]
\centering\footnotesize
\resizebox{\columnwidth}{!}{
\small
\begin{tabular}{c|ccccccc}
\toprule
& pSp & e4e & ReStyle & HS & HFGI & StyleRes & \textbf{Ours} \\
\midrule
Smile & 21.66 & 21.51 & 13.87 & 15.65 & 15.17 & 17.03 & \textbf{14.50} \\
Gender & 26.45 & 29.31 & 26.14 & 19.92 & 19.24 & 19.90 & \textbf{19.46} \\
Lipstick & 40.88 & 40.27 & 34.27 & 33.45 & 31.30 & \textbf{28.53} & 31.21 \\
\midrule
Average $\downarrow$ & 29.66 & 30.36 & 24.76 & 23.01 & 21.91 & 21.82 & \textbf{21.72} \\
\bottomrule
\end{tabular}
}
\vspace{-0.3cm}
\caption{Quantitative comparison on editability.}
\label{tab:editabiliy}
\end{table}

\noindent Recently, StyleRes \cite{pehlivan2023styleres} proposed a method to quantitatively measure editability using FID (Frechet Inception Distance), based on image distributions from the CelebA-HQ annotation dataset.
Specifically, after selecting a feature to edit, the Inception network output distribution of real images that possess the desired feature (positive) is computed.
Then, real images that do not have the feature (negative) are edited, and the Inception network output distribution of the resulting fake images is obtained.
The underlying idea is that the more realistic the edited images are and the better the feature is reflected, the smaller the distance between these two distributions will be.
In Table \ref{tab:editabiliy}, we compared FID editability related to three features, \eg, smile, gender, and lipstick.
In two features out of three, ours showed the lowest FID and the second lowest in the rest feature.
Overall, we compared the average FID, where ours showed the lowest score out of every baseline.

\section{Ablation Studies}
\label{sec:wine_design}

\begin{figure*}[!t]
\begin{center}
\includegraphics[width=0.86\textwidth]{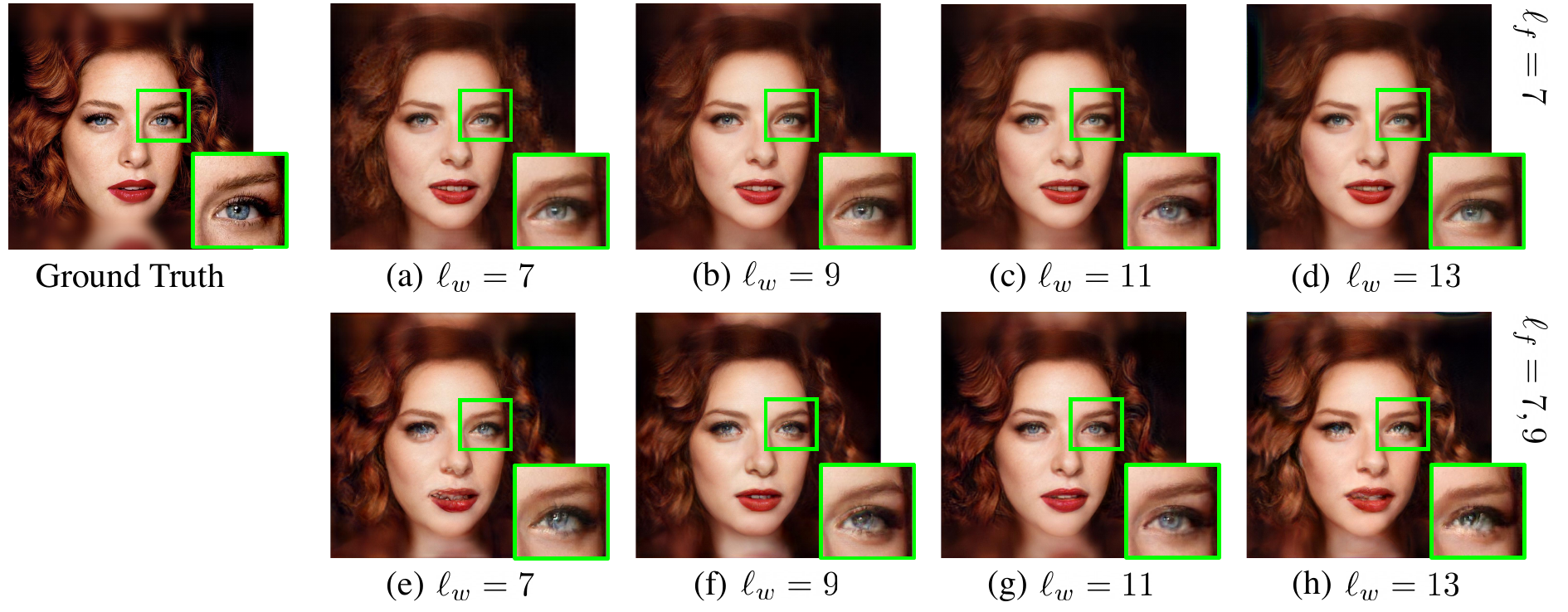}
\end{center}
\caption{
\textbf{Qualitative Comparison of WINE Inversion with Fusion in Different Layers.} Each image represents the inversion results for each scenario in Table \ref{tab:ablation_layer}. The first row (a)-(d) displays inverted images with feature fusion in a single layer $\ell_f=7$, with wavelet fusion in layer $\ell_w=7$,  $\ell_w=9$, $\ell_w=11$, and $\ell_w=13$, respectively. The second row (e)-(h) displays inverted images with feature fusion in multi-layers $\ell_f=7$ and $9$, with wavelet fusion in layer $\ell_w=7$,  $\ell_w=9$, $\ell_w=11$, and  $\ell_w=13$, respectively. We recommend you zoom in for a careful look into the details.}

\label{fig:ablationlayer}
\end{figure*}

\subsection{Choice of Fusion Layer}
We additionally provide both quantitative and qualitative ablation results for the inversion performance of WINE with fusion in different layers. Note that in our main experiment, we apply feature fusion in layers $\ell_{f}=7$ and $9$, and wavelet fusion in layer $\ell_{w}=11$. Each layer corresponds to a fusion of spatial features with resolution $64 \times 64$ and $128 \times 128$, and wavelet coefficients of dimension $w \in \mathbb{R}^{12\times 128 \times 128}.$ 

From the quantitative results in Table \ref{tab:ablation_layer}, we observed that the feature fusion on two layers $\ell_{f}=7$ and $9$ showed better reconstruction accuracy than on a single layer $\ell_{f}=7$. Additionally, wavelet fusion in lower layers ($\ell_{w}<11$) was not sufficient enough to preserve the high-fidelity details, especially in the high-frequency region \ie $\mathcal{L}_{wave}$. Wavelet fusion in the higher layer ($\ell_{w}=13$) also degraded the inversion performance, which can be more carefully observed in Figure \ref{fig:ablationlayer}.

Figure \ref{fig:ablationlayer} shows the inverted images for each scenario in Table \ref{tab:ablation_layer}. It is noticeable that fusion in a single layer (a)-(d) failed to retain high-frequency details like the hand and hair texture. Comparably, in the case of multi-layer feature fusion (e)-(h), inverted images reconstructed more high-frequency details. Yet, wavelet fusion in the lower layers (e), (g), and higher layers (h) generated unwanted distortions, which eventually degraded the image fidelity. Overall, our scenario (g) empirically showed the most promising reconstruction quality, generating realistic images with the least distortion.

\begin{table*}[!t]
\caption{\textbf{Ablation of the Fusion Layers for WINE.}
We compared the inversion performance of WINE with feature and wavelet fusion in different layers. Feature fusion on layers $\ell_f=7$ and $9$, and wavelet fusion on layer $\ell_w=11$ consistently showed the highest fidelity and reconstruction quality among all scenarios.
}
\small
\centering
\begin{tabular}{cc|ccccc}
\toprule
Feature Fusion & Wavelet Fusion  & $L_{2} \downarrow$  & $L_{wave}\downarrow$  & LPIPS $\downarrow$ & SSIM $\uparrow$ & ID sim $\uparrow$ \\
\midrule
\multirow{4}{*}{$\ell_{f}=7$} &  $\ell_{w}=7$  & 0.028 & 0.359 & 0.365 & 0.667 & 0.796\\
  & $\ell_{w}=9$ & 0.026 & 0.356 & 0.362 & 0.701 & 0.830 \\
 & $\ell_{w}=11$  & 0.026 & 0.325 & 0.364 & 0.727 & 0.847 \\
  & $\ell_{w}=13$ & 0.024 & 0.314 & 0.366 & 0.727 & 0.845 \\
\midrule
\multirow{4}{*}{$\ell_{f}=7$ and $9$} & $\ell_{w}=7$  & 0.020 & 0.327 & 0.346 & 0.711 & 0.849 \\
  & $\ell_{w}=9$ & 0.016 & 0.289 & 0.330 & 0.724 & 0.880 \\
 &  $\ell_{w}=11$  & \textbf{0.011}  & \textbf{0.230} & \textbf{0.277} & \textbf{0.753}  & \textbf{0.906} \\
  & $\ell_{w}=13$ & 0.020 & 0.307 & 0.342 & 0.722 & 0.861 \\

\bottomrule
\end{tabular}
\label{tab:ablation_layer}
\end{table*}

\begin{table*}[!t]
\caption{\textbf{Ablation of the Fusion Methods for WINE.}
We compared the inversion performance of WINE with the model which uses wavelet fusion instead of feature fusion.
Though changing all the fusion methods with the feature fusion achieves better results than HFGI, still it shows a big performance degradation compared to WINE.}
\small
\centering
\begin{tabular}{cc|cccc}
\toprule
Model & Fusion Layers & $L_{2} \downarrow$  & $L_{wave}\downarrow$  & SSIM $\uparrow$ & ID sim $\uparrow$ \\
\midrule
\multirow{2}{*}{HFGI}  & $\ell_f=7$  & 0.023 & 0.351  & 0.661 & 0.864\\
  & $\ell_f=7, 9, 11$ & 0.036 & 0.377 & 0.704 & 0.795\\
\midrule
\multirow{2}{*}{WINE} & $\ell_f=7, 9, 11$ & 0.017 & 0.302 & 0.699 & 0.873\\
 & $\ell_f=7, 9$ and $\ell_w=11$ & \textbf{0.011} & \textbf{0.230} & \textbf{0.753} & \textbf{0.906} \\
\bottomrule
\end{tabular}

\label{tab:ablation_hfgi}
\end{table*}

\subsection{Design of Fusion Methods}
To prove the effectiveness of the wavelet fusion, we compared the performance of WINE with the model which uses the feature fusion, proposed in HFGI \cite{wang2022high}, instead of the wavelet fusion in the same resolution layer.
In Table \ref{tab:ablation_hfgi}, we compared the performance of models with the following four settings: The original HFGI which uses feature fusion at $\ell_{f}=7$, HFGI with additional feature fusion at $\ell_{f}=9$ and $11$, WINE with the feature fusion at $\ell_{f}=7$, $9$, and $11$, and the original WINE which uses the feature fusion at $\ell_{f}=7$ and $9$, and the wavelet fusion at $\ell_{w}=11$.
First, simply adding the feature fusion to the higher layer is not helpful for improving the model.
If we change it to the WINE method, \ie, change the generator and add the wavelet loss, the performance significantly improves.
After changing the feature fusion at the $11^{th}$ layer, the performance remarkably improved and achieved state-of-the-art results on various metrics.

\section{Limitation and Future Work}

\noindent Our proposed WINE excels in producing high-quality images by efficiently transferring the residual high-frequency information to the generator. However, we only provided empirical results with specifically SWAGAN, a wavelet-based StyleGAN as the generator.
As our proposed wavelet fusion pertains to generators with intermediate wavelet coefficients, we can potentially generalize our approach to other wavelet-based generators that provides inversion and editing abilities. We will work on applying our method to the inversion of wavelet-based diffusion models and other wavelet expanded generators in the near future.

\end{document}